\documentclass[12pt,a4paper,twoside]{report}

\usepackage[vmargin=20mm,hmargin=25mm]{geometry}  
\usepackage{graphicx} 
\usepackage{parskip}  
\usepackage{setspace} 
\usepackage{refcount} 
\usepackage{upquote}  
\usepackage{soul,color,xcolor}

\usepackage{multirow}
\usepackage{amsthm}
\usepackage{amsmath}
\usepackage{blindtext} 
\usepackage{booktabs}
\usepackage{tabularx}
\usepackage{comment}
\usepackage{longtable}
\usepackage{hyperref}

\newtheorem{example}{Example}[chapter] 

\usepackage[numbers,compress,sort]{natbib}

\newif\ifsubmission 

\title{Generation of Explanations for Logic Reasoning}
\author{Yanyi Pu}
\date{June 2023}
\newcommand{\candidatenumber}{9670K}
\newcommand{\college}{St Edmund's College}
\newcommand{\course}{Master of Philosophy in Advanced Computer Science}

\begin{document}

\begin{sffamily} 

\begin{titlepage}
\makeatletter

\hspace*{-14mm}\includegraphics[width=65mm]{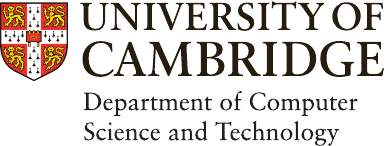}

\ifsubmission

\begin{Large}
\vspace{20mm}
Research project report title page

\vspace{35mm}
Candidate \candidatenumber

\vspace{42mm}
\textsl{``\@title''}

\end{Large}

\else

\begin{center}
\Huge
\vspace{\fill}

\@title
\vspace{\fill}

\@author
\vspace{10mm}

\Large
\college
\vspace{\fill}

\@date
\vspace{\fill}

\end{center}

\fi

\vspace{\fill}
\begin{center}
Submitted in partial fulfilment of the requirements for the\\
\course
\end{center}

\makeatother
\end{titlepage}

\newpage









\end{sffamily}

\begin{figure}[ht]
\centering
\includegraphics[width=1\textwidth]{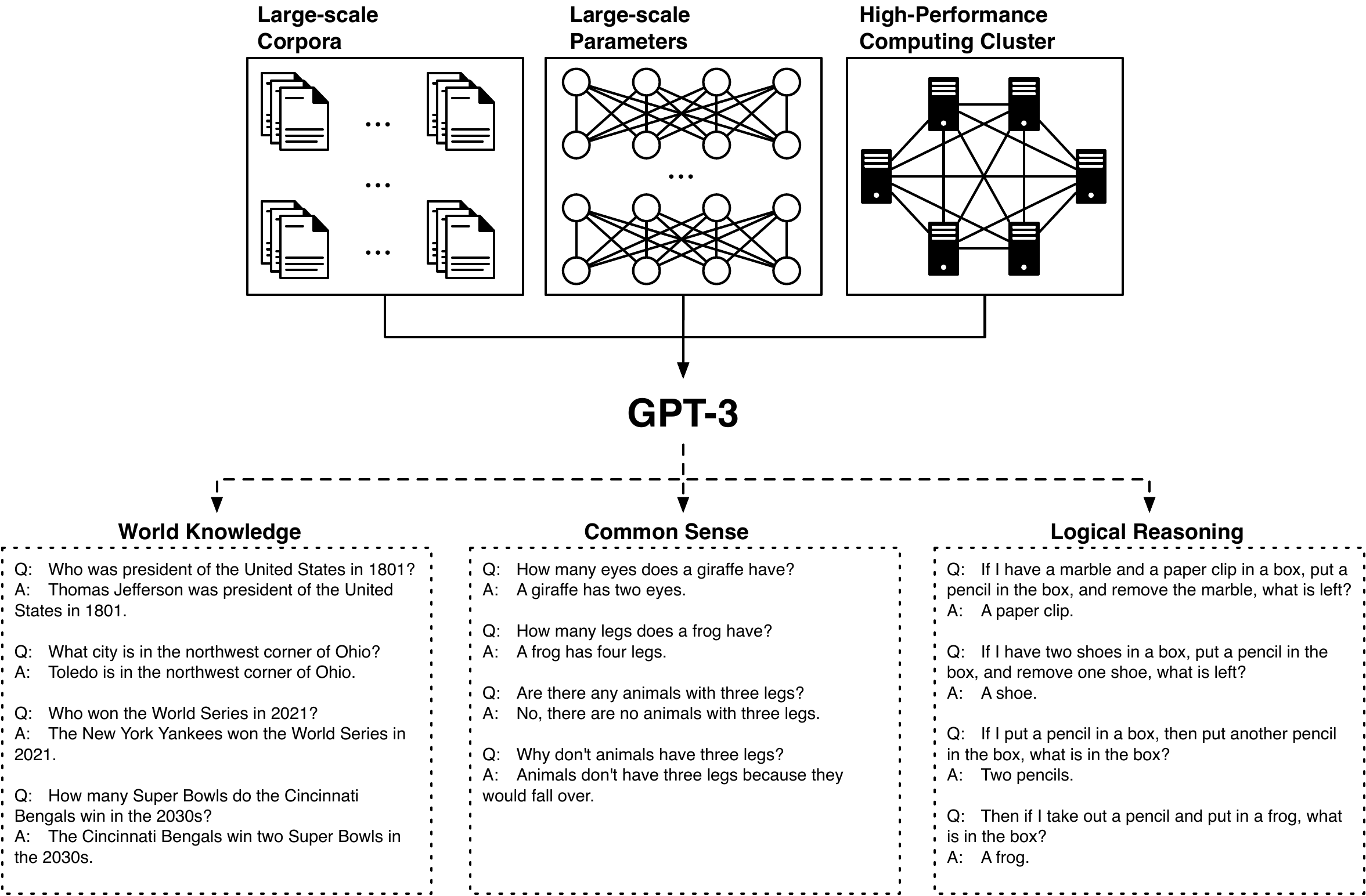}
\caption{GPT-3 has shown the ability of learning world knowledge, common sense, and logical reasoning \cite{han2021pretrained}.}
\label{fig:gpt3}
\end{figure}

\vspace{\fill}
\onehalfspacing
\ifsubmission\else\makeatletter
\textbf{\Huge Declaration}
\vspace{40pt}

I, \@author\ of \college, being a candidate for the \course, hereby
declare that this report and the work described in it are my own work,
unaided except as may be specified below, and that the report does not
contain material that has already been used to any substantial extent
for a comparable purpose.

Figure~\ref{fig:transformers} in Chapter~\ref{Chapter:related_work} is the work of \citet{han2021pretrained}\\
Figure~\ref{fig:pretraining_techniques} in Chapter~\ref{Chapter:related_work} is the work of \citet{ding2022delta}\\
Figure~\ref{fig:gpt3} in Chapter~\ref{Chapter:related_work} is the work of \citet{han2021pretrained}\\
Figure~\ref{fig:chatgpt} in Chapter~\ref{Chapter:related_work} is the work of \citet{zhou2023comprehensive_PFMs}

\bigskip 
\textbf{Signed: Yanyi Pu}

\bigskip
\textbf{Date: 1-June-2023}
\vspace{\fill}
\makeatother\fi

\chapter*{Abstract}

This thesis embarks on an unprecedented exploration into the complex field of \textit{a fortiori} arguments, a critical form of deductive reasoning that draws conclusions upon comparison. It asserts that if one proposition holds true, it must also hold true in a more (or less) extreme case by its relative likelihood. Given its wide-ranging implications in law, philosophy, and artificial intelligence, the \textit{a fortiori} argument highlights the pressing need for in-depth research. This research aims to automate the analysis of such arguments utilizing the advanced language model GPT-3.5-turbo. The experimental approach is thoughtfully devised, aiming to comprehend the underlying reasoning process, generate lucid and logically consistent explanations, and produce similar or novel arguments for subsequent studies, all bolstered by tailored prompting techniques.

The GPT-3.5-turbo is configured to sequentially undertake diverse tasks within a comprehensive pipeline, encompassing reasoning, interpreting, and augmenting \textit{a fortiori} argument. The intermediate reasoning steps facilitated by chain-of-thought aid in comprehending the \textit{a fortiori} arguments, which involve identifying such arguments among `let alone' sentences, discerning the two comparative cases (termed {\em correlate} and {\em remnant}), and categorizing arguments by their relational types and logical categories. The interpretation phase aims to predict hidden properties, integrating them with other information to craft final explanations. Experiments in this phase were executed under two conditions: 1) autonomous model reasoning and interpretation; 2) providing external information at every step to enhance argument understanding and interpretation. The augmentation phase initiates with a detailed analysis of the current argument, followed by the creation of semantically aligned or new instances.

Experimental results indicate that GPT-3.5-turbo sometimes grapples with the intricate details of \textit{a fortiori} arguments, leading to challenges in their detection and classification. When extracting the {\em correlate} and {\em remnant} and predicting hidden properties, the model demonstrates performance comparable to the best model from \citet{razuvayevskaya_2022}'s experiments, which were specifically trained or fine-tuned for these tasks. In most instances, the model can synthesize reasoning results and predicted hidden properties into meaningful and grammatically correct explanations. Meanwhile, evidence from the experiments underlines the efficacy of embedding external information in the reasoning and interpretation processes, leading to significant improvements in generating high-quality explanations. Experiment results also spotlight the model’s superior proficiency in augmenting existing arguments to expand the diversity of the existing dataset. Despite these successes, this research candidly acknowledges limitations and challenges, pointing to valuable directions for future exploration.

This thesis contributes significantly to the fields of artificial intelligence and logical reasoning. The novel methods, rigorous evaluation framework, profound insights, and comprehensive result analyses form a bedrock for future research and open new avenues for exploration in automated logical reasoning \footnote{My MPhil thesis presentation is available at \href{https://drive.google.com/file/d/1wLIBsjfLvO11PjCS6qx4Y9UgRBUfq3wQ/view?usp=sharing}{Thesis Presentation}. I am currently a PhD candidate at the Centre for Doctoral Training in Speech and Language Technologies (CDT in SLT) at the University of Sheffield. For inquiries or further discussion related to the thesis, please contact me via email at ypu17@sheffield.ac.uk.}.

\ifsubmission\else

\chapter*{Acknowledgements}

The journey to complete this thesis has been fraught with unprecedented difficulties and challenges. My time at Cambridge was marked by the end of a long-term relationship, family emergencies, physical injuries, and misunderstandings that tested my mental fortitude. Despite these hardships, I managed to overcome these trials, thanks to the unwavering support and encouragement from many people.

Firstly, I extend my gratitude to my supervisors, Prof. Simone Teufel, Dr. Smaranda Muresan, and especially Dr. Olesya Razuvayevskaya. Their insightful guidance and encouragement have been instrumental in my research journey. Dr. Razuvayevskaya offered endless dedication and support during my darkest times, serving as a beacon of light guiding me through the challenges I faced.

I would also like to acknowledge many academics, including Dr. Weiwei Sun, Prof. Paula Buttery, Prof. Andreas Vlachos, Dr. Petar Veličković, and Prof. Pietro Liò. Their generously shared knowledge and encouragement have been invaluable to my learning and development. Special thanks go to Lise Gough and the administrative team at the Computer Science Department for their consistent assistance.

I am deeply indebted to St Edmund’s College's community, including students, academic fellows, administrative, and maintenance staff. Special appreciation goes to the head tutor, Dr. Judith Bunbury, the student welfare officer, Mr. Isaac Samuel Thor Wilkinson, and the college nurse, Ms. Taryn Rothwell, for their timely and thoughtful care. Additionally, I extend warm thanks to friends at St Edmund’s College, Xiangjian Jiang, Jiachen Liu, Siyu Su, Yuwei Liang, Tao Tang, and to my best friends Qingquan Tian, Lewis Zheng, Yazhe Su, Alex Bi, and many more.

My recovery would not have been possible without the professional care of my GP, Dr. Michael R. Cunningham, my mental health advisor, Mr. Ryan Hegarty, and my Psychiatrist, Dr. Nicola Marshall. I cannot thank them enough for their compassion and expertise.

Last but never least, I owe a profound debt of gratitude to my parents, my strongest source of support and unconditional love. Their belief in me and dedication to my success has been the foundation of all my achievements. Their triumphs are reflected in everything I have accomplished.

\fi
\cleardoublepage 

\tableofcontents

\chapter{Introduction}
\label{firstcontentpage} 

\section{A Brief Introduction to \textit{a fortiori} Arguments}

The \textit{a fortiori} argument, derived from the Latin phrase meaning \textit{`from the stronger (argument)'}\footnote{Source: Merriam-Webster dictionary https://www.merriam-webster.com/dictionary/a\%20fortiori}, is a form of deductive reasoning that uses comparative logic to assert a proposition \cite{sidgwick1916fortiori}. This type of argument suggests that if a statement holds in one situation, it must also hold in another, either a more (positive) or less (negative) extreme situation based on the relative likelihood of the two situations.

These arguments can be categorized into two primary forms \cite{d2017arguing}:
\begin{itemize}
    \item \textit{'a minori ad maius} (from the lesser to the greater): Asserts that if a condition is true in a minor scenario, it should undoubtedly be true in a more extreme scenario.
    \item \textit{'a maiori ad minus} (from the greater to the lesser): Proposes that if a more potent scenario is disregarded, a weaker one should be dismissed with even higher certainty. 
\end{itemize}
Despite their differences, both forms operate under a shared conceptual framework, varying only in the direction of comparison.

The \textit{a fortiori} arguments are not only prevalent in academic discourse but also exert a significant influence on everyday decision-making and persuasive endeavours. These arguments manifest themselves in both straightforward scenarios and intricate societal discussions. As highlighted by \citet{razuvayevskaya_2022}, their prevalence across diverse linguistic contexts underscores the pivotal role they play in logical reasoning and effective communication.

\section{The Reasoning Process of \textit{a fortiori} Arguments}

Reasoning through \textit{a fortiori} arguments entails comparing the relative strengths or likelihoods of two parallel scenarios. This methodological approach begins with the identification of the two compared cases, the correlate (the initially presented scenario), and the remnant (the subsequent scenario). It is crucial to understand that the correlate often forms the premise, typically portraying a hypothetical situation, while the remnant provides the conclusion, typically echoing real-life situations. These components then collectively act as the foundation for comparison, with their relative significance shaping the argument's direction.

One of the inherent challenges in \textit{a fortiori} reasoning is discerning and deducing comparative hidden properties. Even though the scenarios in such arguments might seem similar, they diverge based on certain features like size, magnitude, application, and intensity. These attributes, usually implicit, demand either a broad understanding of the world or domain-specific insights to fathom their relationship with the compared scenarios. These properties are pivotal in \textit{a fortiori} logic as they facilitate case comparison along specific dimensions, reinforcing the core claim. The process of inferring the implicit comparison attributes represents a specific case of enthymeme reconstruction, where implicit logical links are present. Enthymeme reconstruction involves uncovering and adding the missing premises or conclusions, making the argument more overtly logical. Further complexity of {\em a fortiori} interpretation concerns ensuring a clear logical progression and pinpointing the argument's direction, given that various implicit forms of negation often exist without clear linguistic indicators \cite{razuvayevskaya_2022}.

Consider this \textit{a fortiori} example:
\begin{example}
Example:  He could not lift a chair, let alone a sofa\\
1. Premise: He was unable to lift a chair.\\
2. [Unstated premise]: Lifting a sofa requires more effort/physical strength than lifting a chair [can be inferred from universally accepted general knowledge].\\
3. Conclusion: Consequently, he was unable to lift a sofa.\\   
\end{example}

This example aligns with the steps for \textit{a fortiori} reasoning: understanding the comparable actions, identifying underlying properties, understanding the relationship between these properties and the scenarios, gauging the argument's trajectory, and deducing the more probable outcome based on the context.

\section{Automating the Interpretation of \textit{a fortiori} Arguments: Challenges and Research Directions}

While humans can easily grasp \textit{a fortiori} logic, automating such understanding is highly challenging for AI systems. Early efforts were directed towards establishing the argument scheme \cite{kienpointner1992alltagslogik}, specifically for a-fortiori arguments \cite{Sion2013-SIOAFL}. Building on this foundation, \citet{razuvayevskaya_2022} delved into the potential of automating three core tasks in \textit{a fortiori} argument interpretation: argument component detection, relation type classification, and property generation, leveraging recent advancements in NLP. \citet{chen2020generating} further focused on a specific type of {\em a fortiori} arguments to craft explanations involving simple quantity comparison through a symbolic system and simple LSTM.

Despite the pioneering efforts to automate the task of \textit{a fortiori} argument interpretation, achieving the objective generation of the implicit properties and natural language explanations remains a challenging task due to a variety of reasons:

\begin{itemize}
   \item \textbf{Elliptic Semantic Structure}:
   The detection of the correlate and remnant in \textit{a fortiori} arguments relies on a complex semantic phenomenon of ellipsis resolution that modern NLP parsers struggle with. Moreover, the ambiguity of natural language often makes such interpretations challenging even for humans.
   \item \textbf{The Hidden Property}: The insertion of the hidden property that defines the comparison between the correlate and the remnant is an inherently subjective task. 
    \item \textbf{Multidimensional Challenges}: \textit{A fortiori} argument processing spans a series of intricate steps, from textual interpretation to logical explanation crafting. These steps can be broken down further into tasks such as component identification, relationship classification, property prediction, and explanation generation. Given their hierarchical nature, this entire process presents a multidimensional challenge. Developing a system that seamlessly integrates these tasks, while acknowledging their interdependence is a challenging NLP task.
    \item \textbf{Data Scarcity}: Modern NLP models, especially those grounded in deep and reinforcement learning, are heavily dependent on vast annotated datasets. However, there is a significant gap when it comes to datasets for \textit{a fortiori} arguments, particularly when contrasted with prevalent tasks like text classification or named entity recognition. The datasets that do exist are specialized and limited, restricting the reach of data-centric approaches.
    \item \textbf{Domain Independence}: Building a genuinely adaptable pipeline requires a domain-neutral strategy. The ideal system should proficiently decipher \textit{a fortiori} arguments, regardless of their specific theme or domain. Meeting this criterion adds a layer of complexity, as the system should be versatile enough to accommodate a spectrum of topics and linguistic variations.
\end{itemize}

In addition to the challenges of automation, the lack of specialized evaluation tools to assess the quality of the generated explanations poses limitations on the development of the automatic tools aimed at this objective. The endeavour to establish a robust evaluation framework is further complicated by several factors:
\begin{itemize}
    \item \textbf{Diverse Interpretations}: An \textit{a fortiori} argument can often be explained in multiple valid ways. Determining the most apt or accurate explanation is not straightforward. Weighing different explanations often involves a delicate balance between simplicity, clarity, precision, and thoroughness.
    \item \textbf{Nuanced Logic}: The inherent complexity of \textit{a fortiori} arguments exceeds that of their straightforward syllogistic counterparts. These arguments delve deep into comparative analysis, gradations, and layered logical interpretation, making the generation and evaluation of explanations a subjective and challenging task.
    \item \textbf{Domain-Specific Expertise}: Evaluating \textit{a fortiori} explanations often mandates domain-specific knowledge, particularly when arguments are deeply embedded within a specific context. Evaluators without the requisite expertise might struggle with assessing the explanation's validity.
    \item \textbf{Evaluation Subjectivity}: The evaluation spectrum can exhibit subjectivity when entrusted to human evaluators.  Disparities in interpretations may arise from individual differences in background, expertise, and cognitive comprehension.
    \item \textbf{Quality Metrics Challenges}: Establishing definitive quality measures for explanations presents a significant challenge. Ideally, explanations should be clear, concise, accurate, and comprehensive. However, quantifying these attributes can pose difficulties. For example, one person's perception of clarity and conciseness may be considered vague and inadequate by another annotator.
    \item \textbf{Automated Evaluation Limitations}: Beyond the challenges of human assessment, automated methods have their pitfalls. Common metrics like BLEU \cite{papineni-etal-2002-bleu} or ROUGE \cite{ganesan2018rouge}, which juxtapose the generated explanations against a gold standard reference, might not adequately discern logical subtleties since there can be many linguistic ways of generating good-quality explanations.
\end{itemize}

Recent advancements in Large Language Models (LLMs), especially the GPT series \cite{GPTv1,GPTv2,GPTv3,OpenAI2023} and PaLM\cite{chowdhery2022palm}/PaLM-v2\cite{anil2023palm}, offer promising solutions to the challenges of automating \textit{a fortiori} argument interpretation. These models have consistently set benchmarks in various NLP tasks, such as language understanding, reasoning, and content generation \cite{zhao2023survey}. With a broad contextual grasp, they can efficiently process arguments, and their pretraining on extensive datasets equips them with the general knowledge needed for argument interpretation. This foundation is vital for discerning the subtle components of \textit{a fortiori} arguments. Furthermore, their adaptability across a spectrum of tasks and topics positions them as prime candidates for a general \textit{a fortiori} argument processing pipeline.

Despite the aforementioned challenges, the task of automating the interpretation of \textit{a fortiori} arguments holds immense potential. These arguments are common across many domains, from specialized legal documents to everyday conversations. Enhancing NLP systems with the capability to comprehend and formulate such arguments can enrich their overall linguistic proficiency. Such an enhancement can lead to marked improvements in diverse tasks, including question answering, text summarization, and argument generation.

The potential applications are varied across different domains. In the legal domain, a deeper understanding of \textit{a fortiori} arguments can revolutionize document analysis, decision-making, and even the prediction of judicial outcomes. It also empowers legal professionals to craft compelling arguments. For AI systems designed for debates, mastering these arguments can significantly bolster their debating prowess, making such models more effective in academic, legal, and political arenas.

\section{Research Goals and Technical Contributions}
In light of the significant challenges and the success of large language models, this thesis delves into the proficiency of cutting-edge LLMs, with a spotlight on the GPT-3.5-turbo model. It aims to evaluate their ability to interpret \textit{a fortiori} arguments and to generate persuasive explanations. It additionally explored the potential of such models in generating original {\em a fortiori} arguments using prompt learning. It explores the variety of prompts that are created based on a set of widely adopted prompting strategies, evaluating their ability to direct the GPT-3.5-turbo model towards successfully interpreting \textit{a fortiori} arguments. The primary objectives of this thesis are therefore focused on evaluating the capabilities of GPT-3.5-turbo to: (1) accurately perform a chain of \textit{a fortiori} argument interpretation steps; (2) generate logical and structured explanations for such arguments; (3) formulate new persuasive arguments rooted in comprehending the structure of a given argument. To validate the effectiveness of this approach, this thesis proposes a novel evaluation framework which is validated using human annotation.

The main technical contributions of this research can be summarized as follows:

\begin{enumerate}
    \item An in-depth assessment of GPT-3.5-turbo's capabilities has been conducted, focusing on its proficiency in interpreting \textit{a fortiori} arguments, articulating them with precision, and devising innovative arguments. The robustness of these explorations is reinforced by systematic experimentation.
    \item An innovative approach has been developed to guide the GPT model through complex reasoning challenges using a singular, meticulously designed modular prompt. The adaptability of this design allows for the efficient integration of contextual information and supplementary knowledge. Additionally, the prompt's architecture is crafted to facilitate seamless modifications, accommodating diverse tasks and enhancing operational efficiency.
    \item Through comprehensive experimentation, various augmentation strategies have been assessed and validated. Two strategies have been identified as particularly effective in generating semantically similar arguments and crafting novel training instances that resonate with the essence of the original dataset. These effective strategies not only enrich the diversity and scope of the existing dataset but also position the augmented dataset as a valuable resource for future research.
    \item A comprehensive evaluation framework has been introduced, tailored to appraise a wide array of explanations. This pioneering framework harmonizes automated metrics with human assessment, rigorously examining both the intermediate reasoning steps and the integrity of the final explanations.
\end{enumerate}


The rest of this thesis is organised as follows: Chapter 2 presents a discussion of research work related to this thesis. Chapter 3 provides a detailed explanation of the dataset used in this research, the design of the prompts, the experimental settings and implementation, and the data augmentation strategies. Chapter 4 is devoted to presenting and discussing the experimental results. It additionally discusses the limitations of the proposed approaches and technical constraints. Finally, Chapter 5 provides a conclusion to this thesis and discusses potential directions for future research.

\chapter{Related Work and Technical Background} \label{Chapter:related_work}

\section{Razuvayevskaya’s Classification Taxonomy of \textit{a fortiori} Arguments}

\citet{razuvayevskaya_2022}'s work on enthymemes forms the backbone of this project. Building upon her feasibility study in \cite{razuvayevskaya2017finding}, where the identification and expansion of enthymemes involving \textit{a fortiori} arguments were explored, Razuvayevskaya further refined her theoretical framework in her thesis. Drawing from \citet{kienpointner1992alltagslogik}’s methodology, she discerned between \textit{a maiore} and \textit{a minore} reasoning based on the probabilities or likelihoods of two cases. She also observed that the remnant and correlate comparison in \textit{a fortiori} arguments is often explained through a shared hidden property, which can correlate either positively or negatively with the situation. Although several facets of her theoretical model are touched upon elsewhere in this thesis, it's crucial to delve deep into her taxonomy for classifying \textit{a fortiori} arguments. This is pivotal because it directly pertains to this research, where correctly understanding the type of \textit{a fortiori} arguments paves the way for producing accurate explanations for these arguments.

Razuvayevskaya's taxonomy bifurcates \textit{a fortiori} arguments along two axes: principles of likelihood comparison and argument’s logical flow. In this study, I have termed the first as \textit{`sentence type’} and the latter as \textit{`logic category’}. Each axis comprises four primary classes. Notably, the categorization based on principles of likelihood comparison is not mutually exclusive; there are instances where \textit{a fortiori} arguments straddle multiple categories.

The following is a list of sentence types, accompanied by illustrative examples:
\begin{itemize}
    \item \textbf{Precondition (PC)}: it describes when one action (X) must start before or simultaneously for another action (Y) to occur. For X to be a precondition of Y, X is invariably more likely than Y. In Example~\ref{ex:sexism}, the speaker rejects the possibility of trusting sexism by first rejecting its precondition of accepting it.
    \begin{example}
        We can never accept sexism, let alone trust it.
        \label{ex:sexism}
    \end{example}

    \item \textbf{Specificity (SP)}: it refers to situations where one item (the remnant) is a specific case of another (the correlate) or vice versa. The more specific situation is typically rarer and thus less likely. In Example~\ref{ex:uniform}, the additional constraint makes the man wearing a uniform a specific case of a decent man. Therefore, by rejecting the more general case, the speaker also rejects the specific case.
    \begin{example}
        No decent man would do this to a lady, let alone one in uniform!
        \label{ex:uniform}
    \end{example}

    \item \textbf{Quantity(QU)}: This relates to cases where the remnant and correlate represent quantifiable entities that can be measured using standard measurements. The comparison in this category is straightforward: if a million dollars falls short, a lesser amount will undoubtedly be insufficient
    \begin{example}
        Million dollars wouldn’t be enough for the company to get out of its debt let alone two hundred thousand dollars that they have in profit.
    \end{example}

    \item \textbf{Resource Allocation (RE)}: it refers to scenarios where the comparison is based on an implicit, non-standardly measurable resource, such as experience, intelligence, or knowledge. In Example~\ref{ex:attention}, attention and affection are compared by their degree of emotional involvement, which are implicit and could not be quantified by standard measurements.
    \begin{example}
        This isn't a man worthy of our attention, let alone our affection.
        \label{ex:attention}
    \end{example}  
    
\end{itemize}

\citet{razuvayevskaya_2022} categorizes the assertive case as positive and the rejected case as negative \textit{a fortiori} logic. Though the expression of negation in the argument can vary and can be challenging to pinpoint, the classification by logic flow remains intuitive.
\begin{itemize}
    \item \textbf{Positive vs. Negative Logic}: If the correlate is deemed more likely than the remnant after setting aside any negations, the argument is termed `negative'. Otherwise, if the remnant is more likely, it is termed `positive'.
    \item \textbf{Simple vs. Reverse Logic}: Arguments display `simple logic' when the item (either remnant or correlate) with the greater inherent value or quantity is also considered more likely to occur. If the item with a smaller inherent value is presented as more probable, then the argument is said to have `reverse logic'.
\end{itemize}

Given these distinctions, there are four potential combinations for \textit{a fortiori} arguments based on logical flow. Table~\ref{table:logic_categories} provides an illustrative example of each combination.

\begin{table}[ht]
    \centering
    \begin{tabular}{|p{3cm}|p{8.5cm}|p{2.5cm}|}
        \hline
        Logic Category & Example & Sentence Type \\
        \hline
        Negative Simple (NS) & I did not expect to finish in the top three, let alone win the race. & Precondition \\
        \hline
        Negative Reverse (NR) & Just the perception that he'll be pulling in oily shrimp, let alone that it might really happen, can greatly reduce the price he can get, he said. & Resource Allocation \\
        \hline
        Positive Simple (PS) & ANY amount of additional rain -- let alone the amounts that typically fall in a tropical storm -- will be significant & Quantity \\
        \hline
        Positive Reverse (PR) & His contract at Milan expires in June and even the 31-year-old himself has admitted that his whole footballing career, let alone his stay at Milan, could be over. & Specificity \\
        \hline
    \end{tabular}
    \caption{Examples for Different Logic Categories. The associated sentence types are also provided to assist in a clearer comparison.}
    \label{table:logic_categories}
\end{table}

\section{Pre-trained Language Models (PLMs) \& Evolution of the GPT Models}
Pre-trained language models (PLMs) have emerged as pivotal in NLP research, transforming a myriad of applications ranging from rudimentary text classification to intricate language generation and reasoning tasks. The genesis of pre-training is anchored in transfer learning in Computer Vision (CV) \cite{zhuang2020comprehensive}, which subsequently thrived in the NLP domain.

\begin{figure}[ht]
\centering
\includegraphics[width=1\textwidth]{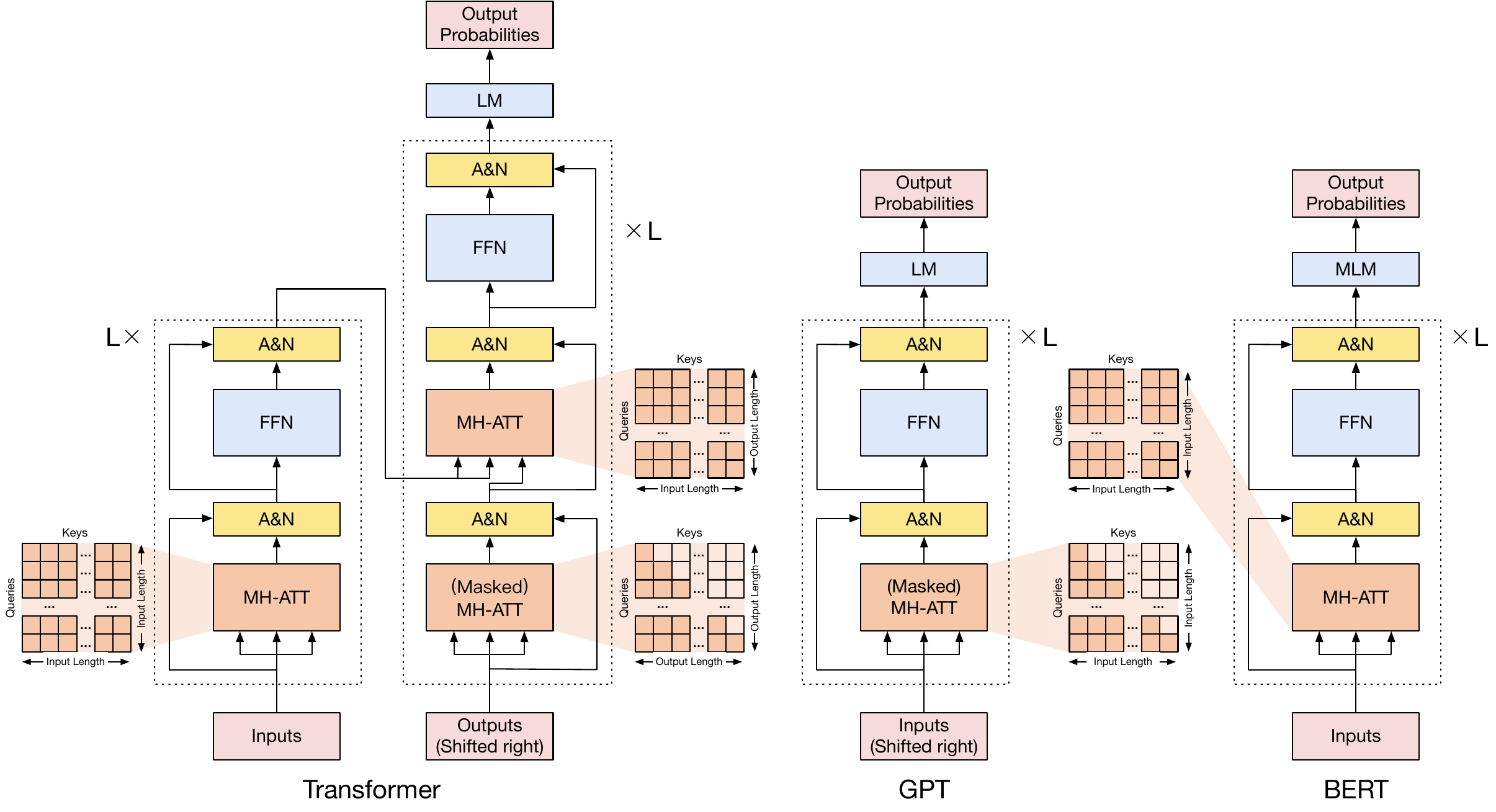}
\caption{The Architecture of Transformer, GPT, and BERT \cite{han2021pretrained}.}
\label{fig:transformers}
\end{figure}

The evolution of PLMs can be chronologically mapped, starting with seminal word embedding techniques such as Word2Vec \cite{word2vec2013} and GloVe \cite{pennington-etal-2014-glove}. These methods pioneered the representation of words in continuous vector spaces, effectively capturing semantic nuances. The introduction of the Transformer architecture by \citet{vaswani2023attention} marked a pivotal turning point in the field. Armed with a self-attention mechanism, the transformers facilitate parallel sequence processing, ensuring efficient training and adept handling of long-term dependencies. Most contemporary PLMs are rooted in this structure, albeit with variations in specific architectures such as using only the Transformer encoder or decoder, or both, as delineated in Figure~\ref{fig:transformers}.

There are three predominant pre-training techniques, as illustrated in Figure \ref{fig:pretraining_techniques}:

\begin{figure}[ht]
\centering
\includegraphics[width=1\textwidth]{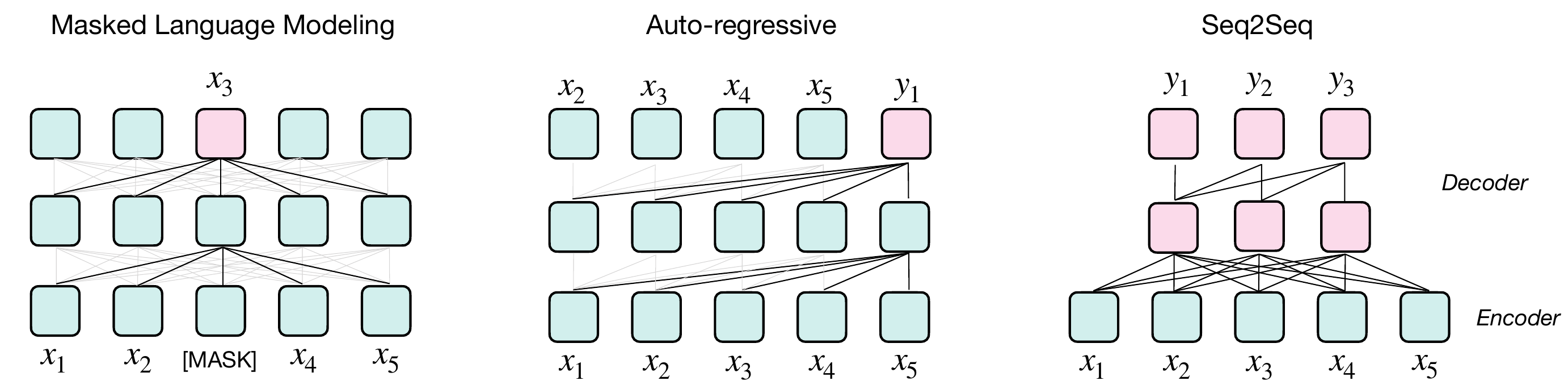}
\caption{Three Predominant Pretraining Paradigms for PLMs \cite{ding2022delta}.}
\label{fig:pretraining_techniques}
\end{figure}

\begin{enumerate}
    \item \textbf{Masked Language Modelling (MLM)}: This involves randomly substituting a certain fraction of the input with special [MASK] tokens, aiming at maximizing the conditional probability of label tokens at [MASK] positions \cite{devlin2019bert}, depicted in Equation~\ref{eq:MLM}, where \(M(x)\) encompasses all masked token locations. Models such as BERT \cite{devlin2019bert} and RoBERTa \cite{liu2019roberta} utilize the Transformer's encoder for pretraining and stand as exemplary instances of this approach. Their bidirectional context-aware representations, combined with pretraining on vast datasets and subsequent fine-tuning, have set a standard solution for diverse NLP tasks.
    \begin{equation} 
    \mathcal{L}_{\mathrm{MLM}}=-\sum_{x_m \in M(\mathbf{x})} \log \mathbf{P}\left(x_m \mid \mathbf{x}_{\backslash M(x)}\right)
    \label{eq:MLM}
    \end{equation}

    \item \textbf{Autoregressive Language Modelling (LM)}: Predicated on the Transformer decoder, its objective is to optimize the probability of the \(i\)-th token conditioned on preceding tokens within a context window, as described in Equation~\ref{eq:ALM}. The unidirectional nature of such models paves the way for efficient training data utilization and superior language generation. The GPT model is recognised as the flagship model championing this pretraining paradigm.
    \begin{equation}
    \mathcal{L}_{\mathrm{LM}}=-\log \mathbf{P}(\mathbf{x})=-\sum_{i=1}^T \log \mathbf{P}\left(x_i \mid x_{i-1}, x_{i-2}, ..., x_{i-k}\right)
    \label{eq:ALM}
    \end{equation}

    \item \textbf{Sequence-to-Sequence Models (Seq2Seq)}: These harness the complete structure of the Transformer, integrating span-level corruption as the primary pre-training task. By randomly masking texts of varying lengths with a solitary [MASK] token, it challenges the model to reconstruct the masked segments. Noteworthy models in this category include T5 \cite{raffel2020exploring_T5}, its successor PaLM \cite{chowdhery2022palm}, and BART \cite{lewis2019bart}. The objective of the Seq2Seq MLM is to maximise the probability of the target sequence when provided with a corrupted sequence.
    \begin{equation}
    \mathcal{L}_{\text {Seq2Seq MLM }}=-\sum_{\mathbf{x}_{i: j} \in M(\mathbf{x})} \sum_{t=i}^j \log \mathbf{P}\left(x_t \mid \mathbf{x}_{\backslash M(\mathbf{x})}, \mathbf{x}_{i: t-1}\right)
    \end{equation}
\end{enumerate}

In this research, the GPT-3.5-turbo \cite{ouyang2022training_gpt3.5} is employed across all experiments. The GPT (Generative Pre-trained Transformer) series models, released by OpenAI, have been instrumental in shaping the field of NLP. Beginning with GPT-1 \cite{GPTv1}, which was pre-trained on the BooksCorpus dataset employing unsupervised generative pretraining and subsequent task-specific fine-tuning, the series has seen exponential advancements. GPT-2 \cite{GPTv2} scaled to 1.5 billion parameters, pre-trained on a more expansive corpus, setting new standards with its ability to generate coherent, diverse texts. It showcased the ability to perform tasks without explicit training (zero-shot), or with a handful of examples (one-shot, few-shot). GPT-3 \cite{GPTv3} was launched in 2020 and it amplified this prowess, boasting over 175 billion parameters, which is an impressive scale-up over its predecessor. This magnitude of scaling has bolstered its few-shot performance, often rivalling or surpassing previous state-of-the-art models or even human performance \cite{GPTv3}. Empirical observations confirm the model's vast repository of world knowledge and common sense \cite{han2021pretrained}, alongside emergent capabilities like mathematical calculations and logical reasoning \cite{wei2022emergent_ability, han2021pretrained}.

\begin{figure}[ht]
\centering
\includegraphics[width=1\textwidth]{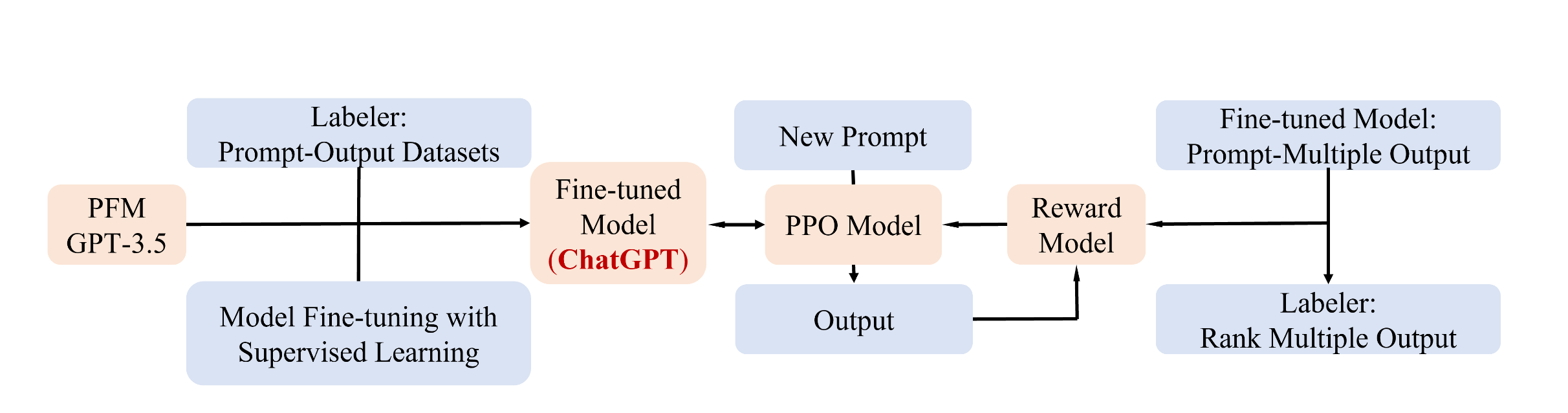}
\caption{ChatGPT is Built on GPT-3.5 and Utilises Reinforcement Learning from Human Feedback (RLHF) \cite{zhou2023comprehensive_PFMs}.}
\label{fig:chatgpt}
\end{figure}

In a significant development in November 2022, OpenAI unveiled ChatGPT, powered by GPT-3.5. This iteration is an enhancement over GPT-3. GPT-3.5 underwent training on a composite of text and code \cite{chen2021evaluating, neelakantan2022text} and was fine-tuned leveraging Reinforcement Learning from Feedback (RLHF) \cite{ouyang2022training_gpt3.5}, as depicted in Figure~\ref{fig:chatgpt}. This strategic modification endowed GPT-3.5 with accelerated and more precise responses to human queries, marking a discernible advancement over GPT-3 in all NLP tasks \cite{zhou2023comprehensive_PFMs}. GPT-4 is the newest advancement in PLMs. While specific technical details are not fully disclosed \cite{OpenAI2023}, it is believed to be magnitudes larger and more advanced than its predecessors \cite{koubaa2023gpt}. Preliminary reports suggest it outstrips GPT-3.5 in reliability, inventiveness, and nuanced instruction comprehension \cite{bahrini2023chatgpt,OpenAI2023,zhou2023comprehensive_PFMs}. On the reasoning front, empirical experiments indicate GPT-4's superior performance over GPT-3.5 in most zero-shot learning reasoning tests \cite{espejel2023gpt35gpt4}. However, both models demonstrate limitations in tasks like Inductive, Mathematical, and Multi-hop Reasoning \cite{espejel2023gpt35gpt4}.

\section{Prompt-based Learning \& Prompting Strategies}

Prompt-based learning has emerged as a strategy to guide pre-trained models to perform specific tasks using carefully designed input prompts and contextual cues, eliminating the need for extensive fine-tuning \cite{liu2021prompting}. Unlike traditional models that are heavily reliant on task-specific training data, prompt-based learning taps into the vast knowledge embedded within large-scale pre-trained models and directs this knowledge using input prompts. The success of GPT-3 has amplified its development, leading to a paradigm shift in NLP from objective engineering to prompt engineering \cite{Sun_2022_paradigmshift}.

\begin{figure}[ht]
\centering
\includegraphics[width=1\textwidth]{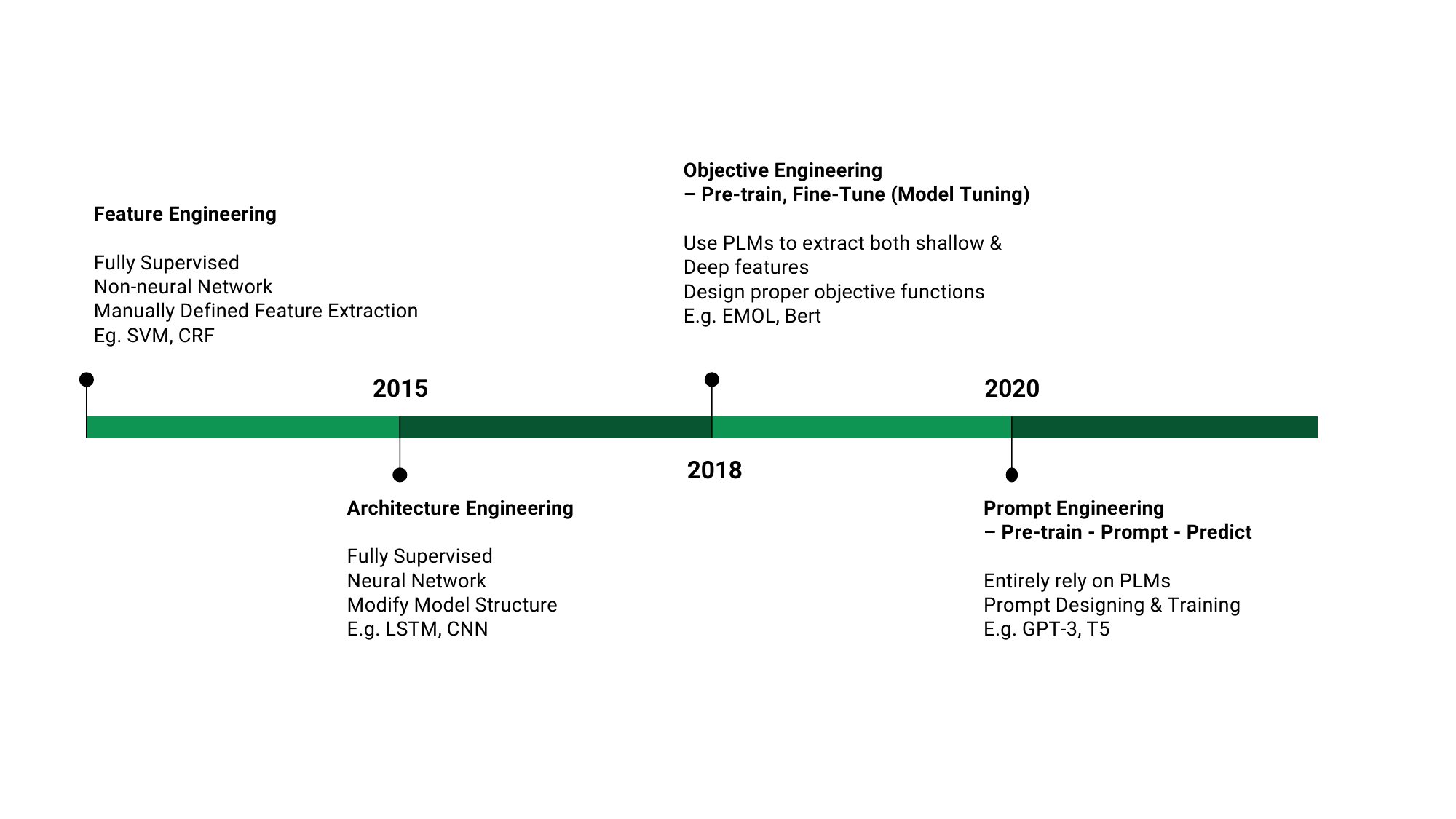}
\caption{Paradigm Shifts in NLP. This figure is a high-level summary of \citet{Sun_2022_paradigmshift}'s survey paper.}
\label{fig:paradigm_shfit}
\end{figure}

This technique has garnered considerable attention due to its exceptional flexibility and adaptability across diverse AI tasks. Employing prompts allows models to generalize from a handful of examples, addressing potential data scarcity issues \cite{schick2020exploiting}. Moreover, it facilitates knowledge transfer across tasks and domains, bolstering the model's adaptability to previously unseen tasks \cite{liu2021prompting}.

A typical prompt comprises several elements \cite{Saravia_Prompt_Engineering_Guide_2022}:

\begin{itemize}
    \item The instruction that specifies the task and its constraints.
    \item The context that provides supplementary information or external knowledge to guide the model towards a desired response.
    \item The input data, which can take various forms.
    \item The output indicator, which governs the type and format of the response
\end{itemize}

Large-scale Language Models (LLMs) can address many NLP tasks in a zero-shot setting. Their performance can be further enhanced using techniques like instruction tuning \cite{wei2022finetuned}, RLHF \cite{christiano2023deep}, or few-shot learning where several examples guide the model towards better performance. However, for complex NLP tasks, LLMs require advanced guidance.

Given these requirements, prompting strategies, often termed \textit{Prompt Engineering}, have emerged. Chain-of-Thought (CoT) prompting facilitates complex reasoning through intermediate steps and has proven effective for intricate reasoning tasks \cite{wei2023chainofthought}. A more recent approach, Tree of Thoughts (ToT), builds on CoT by fostering exploration and enabling LLMs to self-evaluate the progress of intermediate thoughts through a structured reasoning process \cite{yao2023tree}. However, CoT methods are reliant on a fixed set of human-annotated exemplars, and they may not always be effective across varied tasks. In response, \citet{diao2023active} introduced Active-Prompt, adapting LLMs to different task-specific prompts to enhance question-answering capabilities.

LLMs might not possess the comprehensive world knowledge necessary for reliable outputs. This can result in the 'hallucination' issue, where the generated responses sound credible but could be factually wrong or not rooted in the given input \cite{ji2023survey}. To address this, Liu \cite{liu2022generated} integrated external knowledge into prompts, achieving significant performance improvements in common sense reasoning tasks. Similarly, Meta AI proposed Retrieval Augmented Generation (RAG) to tackle knowledge-intensive tasks \cite{lewis2021retrievalaugmented}. RAG amalgamates information retrieval with text generation, allowing for efficient fine-tuning and knowledge modification without retraining the entire model.

Research is also underway to automate prompt optimization for varied tasks \cite{zhou2023large, shin2020autoprompt}. Unlike traditional text-based prompts, \textit{`Soft Prompts'} are non-human-readable continuous parameters, often several magnitudes smaller than LLMs \cite{li2021prefixtuning}. They can be fine-tuned for various NLP tasks using back-propagation on a limited dataset and then utilized as prompts to guide LLMs towards enhanced performance \cite{lester2021power}. This approach offers a cost-effective alternative to traditional fine-tuning methods that adjust the entire set of parameters of the language model \cite{lester2021power}.

\chapter{Dataset and Experiment}

\section{Notation and Problem Formulation}
Generating logical explanations through Language Model Prompting (LM Prompting) involves a sequence of tasks within the Natural Language Processing (NLP) pipeline. This starts by systematically processing the given argument and its associated context. The aim is to accurately grasp both the explicit and implied meanings of the text to understand its underlying logic. Based on this understanding, inferences are drawn in line with the logical progression. The final task is to produce clear and coherent explanations that reflect the primary findings from the previous stages. For \textit{a fortiori} logic, a pivotal aspect is understanding the relationships between correlates and remnants, which often hinge on implied comparative properties and general knowledge. The problem of generating explanations for \textit{a fortiori} arguments can be formalized mathematically as follows:

Given the sentence \(\mathcal{S}\), its surrounding context \(\mathcal{C}\), external information \(\mathcal{I}\), prompt \(\mathcal{T}\), parameterized probabilistic model \(\mathcal{P}_{LM}\), and its window size \(\mathcal{W}\), the goal is to maximize the likelihood of the explanations \(\mathcal{A}\).

\begin{equation}
p(\mathcal{A} \mid \mathcal{S},(\mathcal{C}),(\mathcal{I}), \mathcal{T})=\prod_{i=1}^{|\mathcal{A}|} P_{\mathrm{LM}}\left(a_i \mid \mathcal{S},(\mathcal{C}),(\mathcal{I}), \mathcal{T}, a_{<i}\right)
\label{overal_opti}
\end{equation}

where \(a_i\) represents the \(i\)-th token of the explanation. The components \(\mathcal{C}\) and \(\mathcal{I}\) may be optional. This objective function is subject to constraints that:
\begin{itemize}
    \item Each component in the optimization function, \(\mathcal{S}, \mathcal{A}, \mathcal{C}, \mathcal{I}, \mathcal{T}\), can be expressed in forms compatible with the language model (usually textual strings).
    \item The aggregate length of inputs and outputs does not surpass the language model's window size: \(|\mathcal{S}| + |\mathcal{A}| + |\mathcal{C}| + |\mathcal{T}| \leq |\mathcal{W}|\).
\end{itemize}
In few-shot prompting, \(\mathcal{T}\) comprises \(N\) exemplars of the \(\mathcal{S}, \mathcal{A}\) pair. The Chain-of-thought approaches further incorporate intermediate reasoning steps \(\mathcal{R}\) into the prompt \(\mathcal{T}=\left\{(\mathcal{S}_i, \mathcal{R}_i, \mathcal{A}_i)\right\}_{i=1}^{\mathcal{K}}\). Presuming that the optional components \(\mathcal{C}\) and \(\mathcal{I}\) are independent of the reasoning steps, Equation~\ref{overal_opti} can be simplified and restructured as:

\begin{equation}
p(\mathcal{A} \mid \mathcal{S}, \mathcal{T})=p(\mathcal{A} \mid \mathcal{S}, \mathcal{T}, \mathcal{R}) p(\mathcal{R} \mid \mathcal{S}, \mathcal{T})
\label{simplified}
\end{equation}

where $p(\mathcal{R} \mid \mathcal{S}, \mathcal{T})$ and $p(\mathcal{A} \mid \mathcal{S}, \mathcal{T}, \mathcal{R})$ are defined as:

\begin{equation}
\begin{aligned}
p(\mathcal{R} \mid \mathcal{S}, \mathcal{T}) & =\prod_{i=1}^{|\mathcal{R}|} p_{\mathrm{LM}}\left(r_i \mid \mathcal{S}, \mathcal{T}, r_{<i}\right) \\
p(\mathcal{A} \mid \mathcal{S}, \mathcal{T}, \mathcal{R}) & =\prod_{j=1}^{|\mathcal{A}|} p_{\mathrm{LM}}\left(a_j \mid \mathcal{S}, \mathcal{T}, \mathcal{R}, a_{<j}\right)
\end{aligned}
\label{split_prob}
\end{equation}

In this research, each data entry consists of a sentence \(\mathcal{S}\) paired with a carefully validated set of external details \(\mathcal{I}\). This set includes attributes like the correlate and remnant extracted from the given argument, the relation type holding between them,
and the direction of reasoning. Such information aids in decoding the \textit{a fortiori} argument within the sentence. Depending on the experiment's design, this external information can be integrated into the reasoning process. The modular prompt, as further detailed in Section~\ref{sec:prompt_design}, includes representations of key concepts, reasoning pathways (with optional ground truths for each step), and five sample pairs of the \((\mathcal{S}, \mathcal{A})\) format.

General knowledge is essential when crafting high-quality explanations for \textit{a fortiori} arguments. This is because these arguments often rest on specific world knowledge or assumptions that must be recognized and accepted for the argument to make sense. Typically, this commonsense information is acquired during a language model's pre-training stage and is embedded within its parameters. While users do not have direct control over the exact knowledge the model gains, existing evidence \cite{reynolds2021prompt} suggests that well-crafted prompts can effectively tap into the model's stored knowledge for reasoning tasks.

\section{Dataset}\label{sec:dataset}

The \textit{a fortiori} argument corpus, assembled by \citet{razuvayevskaya_2022}, consists of 3,735 instances. These arguments were harvested from five diverse sources: the British National Corpus \cite{BNCConsortium2007}, English Gigaword v.5 \cite{Parker2011}, Twitter, The Free Library \cite{FreeLibrary}, and web search results for dictionary examples. These sources encompass a vast array of genres and topics, spanning both high and low-register domains. For the purposes of this study, a subset of 1,030 \textit{'let alone'} sentences---referred to as \textbf{`the dataset'} henceforth---has been utilized. This subset was selected based on its cross-annotated nature and human agreement, ensuring both precision and reliability.

\begin{table}[ht]
\centering
\begin{tabular}{|l|p{10cm}|}
\hline
\textbf{Column} & \textbf{Description} \\
\hline
\texttt{text} & The actual sentence text. \\
\hline
\texttt{cor\_start}, \texttt{cor\_end} & The start and end indices of the part of the sentence serving as a correlate in the argument. \\
\hline
\texttt{rem\_start}, \texttt{rem\_end} & The start and end indices of the part of the sentence serving as a remnant in the argument. \\
\hline
\texttt{NAF} & A binary (Yes/No) value indicating whether the sentence is likely or not to contain an \textit{a fortiori} argument. \\
\hline
\texttt{prop1}, \texttt{prop2} & The hidden properties manually annotated in the sentence. \\
\hline
\texttt{logic} & The type of logic used in the sentence (e.g., NS, NR, PS, PR, Undefined). \\
\hline
\texttt{class} & The class or type of sentence (e.g., QU, RE, SP, PC, Undefined). \\
\hline
\texttt{metaphor} & A binary value indicating whether the sentence contains a metaphor. \\
\hline
\texttt{additive} & A binary value indicating whether the sentence is additive. \\
\hline
\texttt{comment} & Comments or additional notes on the sentence. \\
\hline
\end{tabular}
\caption{The Overview of the Curated Dataset.}
\label{tab:dataset_overview}
\end{table}

The dataset is organized tabularly, with 13 distinct columns. Table~\ref{tab:dataset_overview} provides a concise summary. Each argument entry is meticulously indexed, detailing the positions of both correlate and remnant. The dataset also captures hidden properties, the argument's logical structure, sentence type, and binary indicators for metaphors and additive structures. The depth of annotation, achieved through cross-annotation, enhances comprehension of the arguments' subtle differences, proving invaluable for training generative language models.

\begin{figure}[h]
\centering
\includegraphics[width=0.6\textwidth]{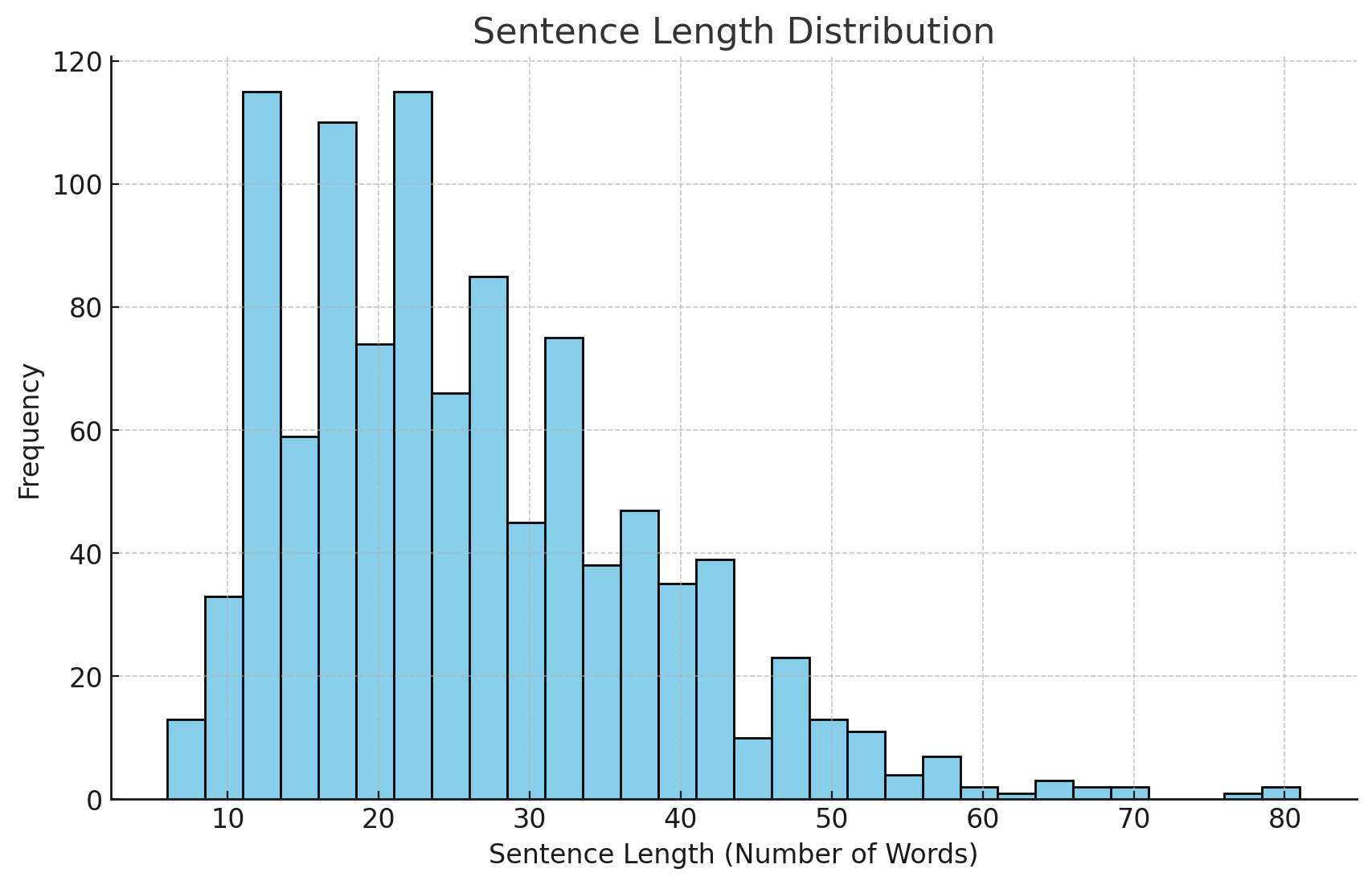}
\caption{Distribution of Sentence Lengths in the Dataset.}
\label{fig:sent_length}
\end{figure}

The sentences in the dataset have an average length of 26 words, ranging from six-word sentences to 81-word ones, as shown in Figure~\ref{fig:sent_length}. Notably, 6.2\% (64/1030) of the sentences, despite employing the \textit{'let alone'} phrase, are not categorized as \textit{a fortiori}. For arguments labelled as 'NAF' with other annotations specified, it suggests that the correlate and remnant positions would need to be swapped to provide coherent interpretation.

\begin{table}[ht]
\centering
\begin{tabular}{clllllll}
\multicolumn{1}{l}{}                                 &                                         & \multicolumn{5}{c}{\textbf{Class}}                                                              &                                     \\ \cline{2-8} 
\multicolumn{1}{l|}{}                                & \multicolumn{1}{l|}{}                   & \textbf{RE} & \textbf{PC} & \textbf{QU} & \textbf{SP} & \multicolumn{1}{l|}{\textbf{Undefined}} & \multicolumn{1}{l|}{\textbf{Total}} \\ \cline{2-8} 
\multicolumn{1}{c|}{\multirow{5}{*}{\textbf{Logic}}} & \multicolumn{1}{l|}{\textbf{NS}}        & 373         & 225         & 20          & 113         & \multicolumn{1}{l|}{0}                  & \multicolumn{1}{l|}{731}            \\
\multicolumn{1}{c|}{}                                & \multicolumn{1}{l|}{\textbf{NR}}        & 40          & 0           & 103         & 1           & \multicolumn{1}{l|}{0}                  & \multicolumn{1}{l|}{144}            \\
\multicolumn{1}{c|}{}                                & \multicolumn{1}{l|}{\textbf{PR}}        & 33          & 10          & 6           & 6           & \multicolumn{1}{l|}{0}                  & \multicolumn{1}{l|}{55}             \\
\multicolumn{1}{c|}{}                                & \multicolumn{1}{l|}{\textbf{PS}}        & 32          & 0           & 18          & 0           & \multicolumn{1}{l|}{0}                  & \multicolumn{1}{l|}{50}             \\
\multicolumn{1}{c|}{}                                & \multicolumn{1}{l|}{\textbf{Undefined}} & 0           & 0           & 0           & 0           & \multicolumn{1}{l|}{50}                 & \multicolumn{1}{l|}{50}             \\ \cline{2-8} 
\multicolumn{1}{l|}{}                                & \multicolumn{1}{l|}{\textbf{Total}}     & 478         & 235         & 147         & 120         & \multicolumn{1}{l|}{50}                 & \multicolumn{1}{l|}{\textbf{1030}}  \\ \cline{2-8} 
\end{tabular}
\caption{Dataset Distribution by Sentence Type and Logic Category.}
\label{tab:dataset_dist}
\end{table}

Table~\ref{tab:dataset_dist} offers a breakdown of the dataset by sentence and logic types, each dimension further bifurcated into four categories and an 'Undefined' category. Key observations from this table are summarised below:

\begin{itemize}
  \item The Resource Allocation (RE) type dominates the sentence categories, followed by Precondition (PC) and Quantity (QU).
  \item Among logic types, Negative Simple (NS) is most prevalent, with Positive Simple (PS) and Positive Reverse (PR) being relatively less frequent.
  \item Sentence categories exhibit considerable diversity in their logic type distribution. For instance, over half of the instances in NS, PS, and PR are linked to Resource Allocation (RE), while a substantial majority (71.5\%) of NR sentences are tied to Quantity (QU).
  \item Sentences classified as undefined in one dimension are likewise undefined in the other, leading to 50 such entries.
\end{itemize}

As per the guidelines set by \citet{razuvayevskaya_2022}, annotators were instructed to select pertinent hidden properties from a predefined list of 23 commonly observed properties in \textit{a fortiori} arguments (refer to Appendix~\ref{Appendix:common_properties}). Additionally, they had the discretion to suggest new properties or leave the fields blank if they felt none were fitting. Consequently, this dataset encompasses 208 unique properties. 88 sentences lack specified properties, typically falling under the Precondition or Specificity types and defaulting to properties like \textit{'typical sequence of actions'} or \textit{'specific case of'}.

\begin{table}[ht]
\centering
\begin{tabular}{clllllll}
\multicolumn{1}{l}{}                                 &                                         & \multicolumn{5}{c}{\textbf{Class}}                                                              &                                     \\ \cline{2-8} 
\multicolumn{1}{l|}{}                                & \multicolumn{1}{l|}{}                   & \textbf{RE} & \textbf{PC} & \textbf{QU} & \textbf{SP} & \multicolumn{1}{l|}{\textbf{Undefined}} & \multicolumn{1}{l|}{\textbf{Total}} \\ \cline{2-8} 
\multicolumn{1}{c|}{\multirow{5}{*}{\textbf{Logic}}} & \multicolumn{1}{l|}{\textbf{NS}}        & 5           & 15          & 5           & 14          & \multicolumn{1}{l|}{0}                  & \multicolumn{1}{l|}{39}             \\
\multicolumn{1}{c|}{}                                & \multicolumn{1}{l|}{\textbf{NR}}        & 5           & 0           & 5           & 1           & \multicolumn{1}{l|}{0}                  & \multicolumn{1}{l|}{11}             \\
\multicolumn{1}{c|}{}                                & \multicolumn{1}{l|}{\textbf{PR}}        & 5           & 5           & 5           & 5           & \multicolumn{1}{l|}{0}                  & \multicolumn{1}{l|}{20}             \\
\multicolumn{1}{c|}{}                                & \multicolumn{1}{l|}{\textbf{PS}}        & 5           & 0           & 5           & 0           & \multicolumn{1}{l|}{0}                  & \multicolumn{1}{l|}{10}             \\
\multicolumn{1}{c|}{}                                & \multicolumn{1}{l|}{\textbf{Undefined}} & 0           & 0           & 0           & 0           & \multicolumn{1}{l|}{20}                 & \multicolumn{1}{l|}{20}             \\ \cline{2-8} 
\multicolumn{1}{l|}{}                                & \multicolumn{1}{l|}{\textbf{Total}}     & 20          & 20          & 20          & 20          & \multicolumn{1}{l|}{20}                 & \multicolumn{1}{l|}{\textbf{100}}   \\ \cline{2-8} 
\end{tabular}
\caption{Distribution for Human Evaluation Set}
\label{tab:human_eva_set}
\end{table}

For manual assessment, 100 sentences were selected from the dataset to form an evaluation set. This set was curated to ensure representation across all sentence types, with each type represented by 20 sentences. To achieve this balance, five sentences were chosen from each combination of sentence type and logic category, totalling 25 combinations. In instances where specific combinations yielded fewer than five sentences, the gap was filled by proportional sampling from other combinations. As shown in Table~\ref{tab:human_eva_set}, the evaluation set maintains balance with respect to the `Class' category, mirrors the 'Logic' distribution from the main dataset, and includes less common cases.

\section{Experiment Design and Implementation}

\subsection{Preliminary Experiment and Model Selection} \label{sec:pre}

The GPT-4 model boasts considerable enhancements in various reasoning tasks when compared with its predecessors \cite{OpenAI2023}. To evaluate these advancements in the context of explaining \textit{a fortiori} arguments, an initial experiment using 12 randomly chosen sentences (3 for each sentence type) was initiated. Both GPT-3.5 and GPT-4 were instructed with producing short and long explanations using an identical prompt\footnote{The general prompt provides the definition of \textit{a fortiori} logic and then solicits explanations for selected sentences. The short explanation is succinct, while the long explanation has no word limit}.

In this preliminary phase, the study sought to accomplish two primary objectives:
\begin{itemize}
    \item Evaluate the default reasoning capabilities of GPT models. 
    \item Assess the performance variations under altered temperature conditions.
\end{itemize}
This dual-focused approach would not only provide an initial performance indicator but also identify the optimal settings tailored for the research problem at hand. The outcomes of this experiment are detailed in Table~\ref{tab:pre_test}.

When instructed to generate short explanations, a consistent trend emerged across models: the explanations largely manifested as paraphrases of the original argument, lacking interpretation capabilities. This is evident from the near-zero scores they collectively registered. In contrast, when evaluating the quality of more extended explanations, there was a discernible performance variation. The GPT-3.5 model, when set to a low temperature, marginally outperformed its default setting with a mean score of 1.25. However, this score paled in comparison to the GPT-4's impressive mean score of 1.83.

A subsequent paired t-test was conducted to ascertain the significance of these observations. The results of the comparison are summarised in Table~\ref{tab:Sig_test}. While short explanations displayed no significant quality differences across models, the long explanations painted a different picture. GPT-4 exhibited a marked enhancement in the quality of long explanations relative to GPT-3.5. Furthermore, even though the paired test between GPT-3.5's default setting and its low-temperature variant did not yield a statistically significant difference, a nuanced examination revealed a slight uptick in performance at the lower temperature.

It's imperative to approach these findings with a measure of caution. The dataset employed for this preliminary test was limited in scope. It may not holistically represent the nuances and diverse characteristics of a more expansive dataset. Potential variations in the significance test outcomes could arise when evaluated against diverse random samples. Moreover, when analysing these results, external factors that might have influenced the outcomes weren't considered, nor was there an exploration into the real-world implications of the observed differences. While the initial findings provide some understanding, they highlight the need for a deeper, more thorough examination in future research stages.

\begin{table}[ht]
\centering
\begin{tabular}{|l|l|l|}
\hline
\textbf{Comparison} & \textbf{p-value (Short)} & \textbf{p-value (Long)} \\ \hline
GPT-4 vs GPT-3.5 default & 0.339 & \(1.3 \times 10^{-4}\) \\
GPT-3.5 low temp. vs GPT-3.5 default & 1.000 & 0.175 \\
GPT-4 vs GPT-3.5 low temp. & 0.298 & 0.048 \\ \hline
\end{tabular}
\caption{Significance Testing for Explanation Quality among Different GPT Models.}
\label{tab:Sig_test}
\end{table}

Given the constraints related to the unavailability of the GPT-4 API at the time of the experiment, the GPT-4 model operated in its default setting. The decision to employ \textit{'gpt-3.5-turbo-16k-0613'} \cite{openai2023gpt35} as the standard model for subsequent experiments, hereafter referred to as 'the GPT-3.5 model', was influenced by the factors such as cost-effectiveness, stability during experimentation, and the ability to handle extensive prompts and demonstrations\footnote{The total prompt length, excluding additional inputs or model outputs, is around 2,500 words. OpenAI's tokenization tool may segment each word into multiple tokens, potentially causing the combined length of inputs and outputs to surpass GPT-3.5-turbo's default window size of 4,096 tokens.}. This model, representing a snapshot of \textit{gpt-3.5-turbo} from June 13th, 2023, was trained on data until September 2021 and will not receive further updates. The utilization of a consistent, non-updating version of \textit{gpt-3.5-turbo} guarantees stable performance across different invocations, crucial for consistent results across varied experiments.

While \textit{gpt-3.5-turbo-16k-0613} parallels the capabilities of the standard \textit{gpt-3.5-turbo} model, it offers an expanded context, four times its original size. Different from other GPT-3.5 variants, this model capitalizes on the Chat Completion API introduced in March 2023, as opposed to traditional text-completion APIs. This chat-based approach has been endorsed for its success in addressing a broad spectrum of conversational requirements, combining enhanced flexibility and precision. Additionally, it stands as a more economical alternative to the models such as Davinci and Ada \cite{openai_pricing}.

\subsection{Fine-tuning Hyperparameters for GPT-3.5-Turbo}

The OpenAI API guide \cite{openai_chatapi} provides a detailed overview of various hyperparameters that users can modify based on their specific needs. Among these, four hyperparameters stood out as potentially influential for generating high-quality explanations for \textit{a fortiori} arguments.

\begin{itemize}
    \item \textbf{Temperature [0,2]\footnote{The square bracket denotes the range of values available for each hyperparameter.} (default = 1)}: This parameter determines the output's randomness. A lower value results in deterministic and concentrated responses, while a higher value brings variability, possibly at the expense of clarity. As per insights from Section~\ref{sec:pre}, lower temperatures are favoured for ensuring logical and comprehensible explanations.
    \item \textbf{Top-P (known as 'nucleus sampling') [0,1] (default = 1)}: Top-P influences output variety by determining the range of probable next tokens. For example, the value of 0.1 would consider only the top 10\% of tokens by probability. Reducing the Top-P value might lead to more focused explanations.
    \item \textbf{Frequency Penalty [-2, 2] (default = 0)}: This parameter penalizes frequent tokens, promoting linguistic diversity while avoiding repetitiveness. A balanced value, close to the default, is optimal for maintaining a blend of both variety and clarity.
    \item \textbf{Presence Penalty [-2, 2] (default = 0)}: This parameter affects tokens based on their current occurrence, fostering a broader token spectrum. A balanced value is preferred to ensure diverse yet coherent explanations.   
\end{itemize}

Following the recommendations of the API guide \cite{openai_chatapi} and in alignment with other research findings, I focused on adjusting the Temperature parameter while maintaining Top-P at its default value of 1. This strategy was adopted to achieve deterministic output, which is of high importance in logical reasoning tasks due to its consistent reliability and enhanced interpretability.

Upon setting the temperature to zero, I observed that the generated output often yielded unsatisfactory explanations. These explanations tended to either rephrase the original statements or follow a recurrent pattern like 'It is more/less likely that…'. Such patterns can be attributed to the model's inclination to produce the 'safest' responses at the minimal temperature setting, emphasizing unwavering consistency and high confidence. However, this conservative approach, while minimizing errors, lacked the depth necessary to produce explanations rooted in the nuanced comparison of underlying properties.

To address this issue, I undertook controlled experiments, incrementing the temperature in steps of 0.05 across a set of 50 randomly chosen sentences and manually evaluating the generated explanations. As the Temperature approached 0.3, the aforementioned concerns were substantially alleviated. As a result, I selected a Temperature of 0.3 for all subsequent experiments. It's crucial to note that this decision is grounded on a limited data subset and primarily targets a specific concern, which means there might be other latent challenges yet to be uncovered.

In terms of frequency and presence penalties, despite experimenting with values within the [-0.5, 0.5] range across multiple runs, no discernible or consistent patterns emerged. Lacking a consensus in the existing literature on how to optimize these parameters, I opted to adhere to their default settings, aiming for a balanced penalty during the output generation.

\subsection{Prompt Design for Logic Reasoning and Explanation} \label{sec:prompt_design}

This section details the design and iterative refinement of a tailored prompt to facilitate the GPT model in generating explanations for \textit{a fortiori} arguments, with an aim to generate desired explanations in a standardized format.

The design journey began with a simple prompt: 
\begin{quote}`Explain the \textit{a fortiori} logic for the given sentence.'\end{quote} Through an output-oriented unit test approach, this prompt underwent refinement. Each iteration involved selecting a random dataset instance, gauging the explanation it produced, and making modifications. The changes comprised of a detailed explanation of the task, guidance, specification of the constraints, output templates, output examples, and the output format. This iterative method not only ensured the prompt's effectiveness but also optimized costs, given OpenAI's token-based costs. The refined prompt was then tested on hundreds of examples to validate its consistency across various comparison relations within {\em a fortiori} arguments.

\begin{figure}[ht]
\centering
\includegraphics[width=1\textwidth]{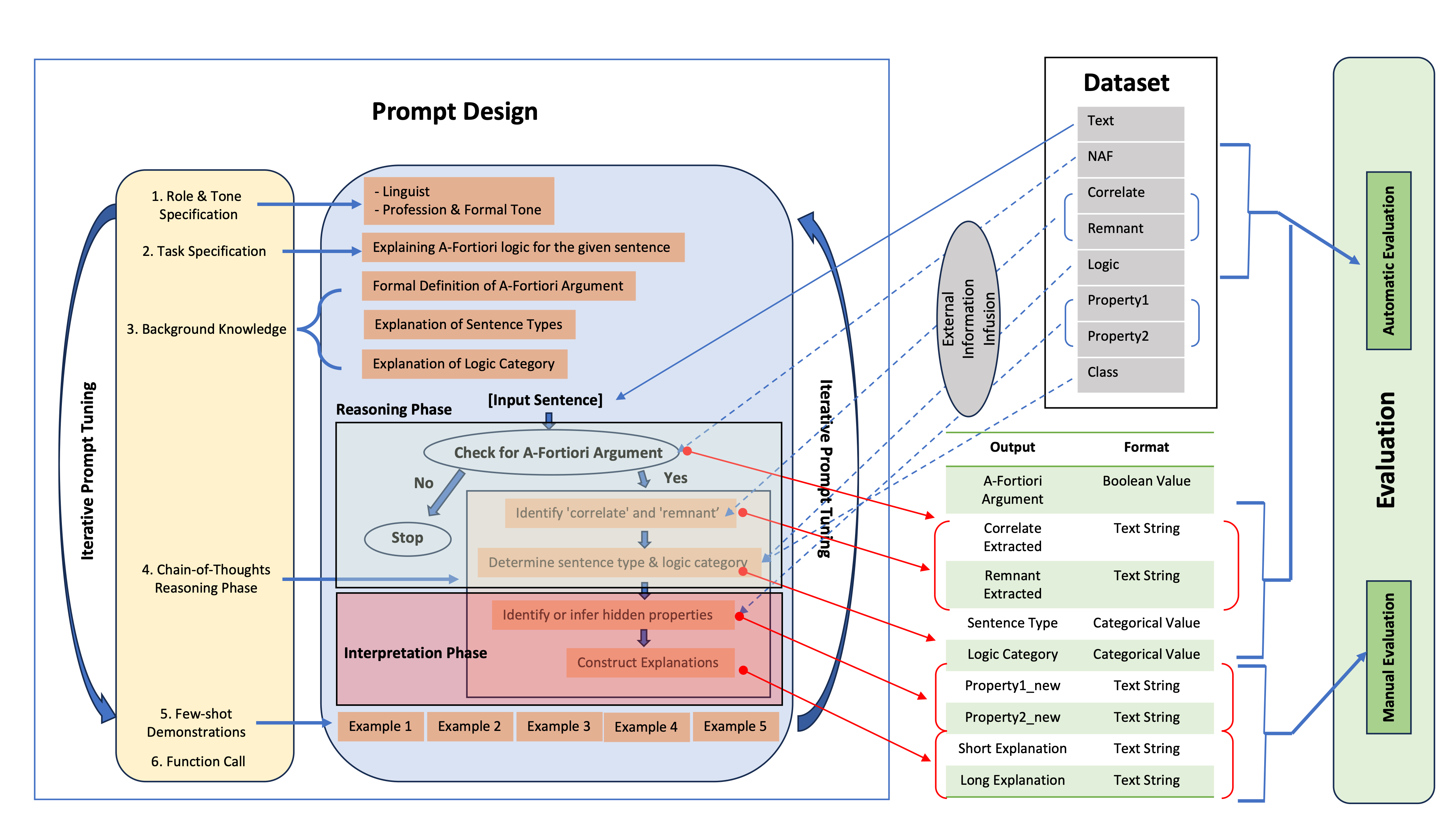}
\caption{This diagram showcases the generation of \textit{a fortiori} argument explanations using prompt learning. The prompt has six main components (in primrose yellow), linked to one or more building blocks (in orange). These blocks provide essential context for understanding the \textit{a fortiori} logic and are integrated via cross-referencing. Each block was optimized iteratively, with the diagram presenting the best placements from experimentation. Blue arrows indicate the model's inputs; dotted arrows represent optional information additions, and the red arrow denotes the model's outputs, some of which are either auto-evaluated against the dataset's ground truth or assessed manually. The prompt's modular design skillfully addresses the intricacies of \textit{a fortiori} argument analysis, facilitates external knowledge integration, and allows adaptability to other tasks with minimal modifications. }
\label{fig:explain}
\end{figure}

The design utilized several prompting techniques including Role \& Tone specification, Task description, Chain-of-thoughts (CoT) reasoning, external knowledge injection, and few-shot learning. 
The prompt's structure is illustrated in Figure~\ref{fig:explain}. Its modular design stands out due to its versatility. Recognizing the complexities of \textit{a fortiori} reasoning, the prompt incorporates rich context, including definitions, concept classifications, key term hints, templates, and examples. For clarity, it's divided into specific sections, such as \textit{`Class'}, \textit{`Logic'}, \textit{`normalize\textunderscore short\textunderscore explanation'}, \textit{`Examples'}, and \textit{`CoT'}. The \textit{`Final\textunderscore prompt'} binds these sections, ensuring cohesion. This structured approach not only supports the model's diverse reasoning capabilities but also ensures easy adaptability to tasks like data augmentation. 

The model assumed the role of a linguist instructed to explain \textit{a fortiori} logic. After being equipped with foundational knowledge of \textit{a fortiori} arguments, sentence types, and logic categories, the model evaluated whether the given sentence truly contained an \textit{a fortiori} argument. If it did not, the reasoning was halted; otherwise, the model proceeded.

The CoT reasoning phase consisted of four major steps:
\begin{enumerate}
    \item Identify the `correlate' and `remnant' terms.
    \item Ascertain the likelihood of the correlate or remnant in context.
    \item Identify or infer two hidden properties affecting the likelihood comparison.
    \item Generate short and long explanations of the \textit{a fortiori} arguments, each with its constraints.
\end{enumerate}

To assist in understanding the argument more accurately, the model was provided with hints and cross references. For instance, the model was reminded that the correlate precedes the discourse marker `let alone', while the remnant follows it; and the logic category can aid in deciphering the likelihood of the comparative terms.

Additionally, the dataset had annotations for the first and third steps, and the mode was provided with the logic category to help  with inferences for the second step. Two scenarios were considered to assess the impact of external knowledge. In the first, the model reasoned autonomously. In the second, ground truths were provided at each reasoning stage. However, it's crucial to note that unlike the teacher-forcing approach, prompt-based learning only suggests the use of external knowledge, allowing the model some discretion.

The quality and format of explanations were controlled by introducing constraints. The model generated two types of explanations: a concise single-sentence version and a more detailed three-sentence version. Both versions required the inclusion of key elements, such as correlate, remnant, and at least one hidden property. For the shorter version, sentence templates were provided to ensure consistency. These templates are detailed in Table~\ref{Appendix:norm_temp}. Few-shot learning was combined with in-context learning in the design by introducing five examples within each prompt along with the surrounding context. A `function\_call' was used to guide the model to produce outputs in the desired format, specifically as JSON objects.

\subsection{Data Augmentation}

 This study is aimed at a challenging reasoning task where the outputs are open-ended. Typically, a comprehensive dataset would be ideal for training a model to tackle such complexity. However, the dataset used here consists of only 1,030 annotated instances, covering 25 combinations of sentence types and logic categories, and some combinations are scarcely represented\footnote{Refer to Section~\ref{sec:dataset} for more details}. Given such limitations, it's challenging for any language model to gain adequate insights.

To counter this limitation, I applied data augmentation techniques to generate additional training instances. The goal was to enrich the dataset's variety and achieve a more balanced distribution. These augmented instances could benefit future research by increasing the quantity and diversity of training samples and enhancing the reasoning process's explainability.

I initially introduced three strategies for data augmentation:

\begin{enumerate}
    \item \textbf{Semantically Similar Sentences}: Generate a sentence resembling the original one, maintaining sentence type, logic category, properties, and keeping the correlate and remnant semantically similar to the ones in the original argument.
    \item \textbf{Reversed Logic Sentences}: Form a new sentence from the original one, keeping the correlate-remnant relation type and properties but reversing the logic category.
    \item \textbf{Novel Sentences}: Generate a completely new sentence based on the original one, retaining its sentence type and logic category but changing the topic and introducing new hidden properties.
\end{enumerate}

During the augmentation experiment, the second strategy, forcing a reversed logic sentence, often produced arguments that sounded unnatural. This might be due to the real-world distribution of logic categories, with NS logic being most prevalent, as shown in the dataset statistics on page. However, Strategies 1 and 3 were proven effective. The first strategy focused on creating instances similar to the originals, while the third introduced more diversity. Though these strategies have the potential to produce countless new instances, I generated only 2,000 new sentences due to time and cost constraints.

Similarly to the interpretation task, data augmentation was executed through prompt-based learning, encompassing the following crucial steps:

\begin{itemize}
    \item First, the GPT model analyzed the input sentence, extracting its underlying sentence and logic types.
    \item Then, The prompt specified the relation type for the new argument, based on the described augmentation strategies.
    \item Finally, the model analyzed the new argument, outputting its correlate, remnant, hidden properties, sentence type, logic category, and associated explanations. It also provided brief topic labels for both the original and augmented sentences.
    \end{itemize}

\begin{figure}[ht]
    \centering
    \includegraphics[width=1\textwidth]{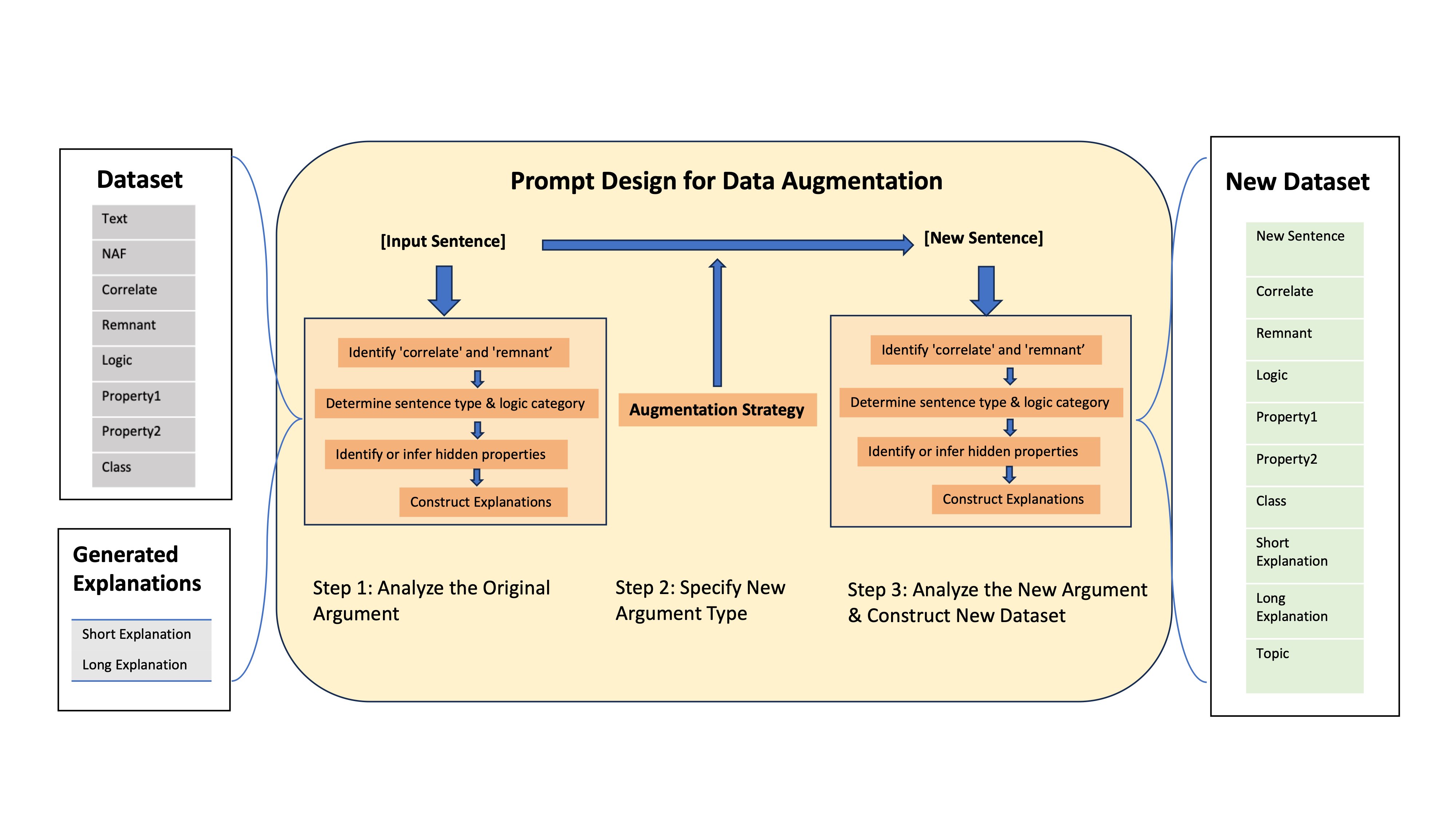}
    \caption{The diagram illustrates the augmentation workflow for \textit{a fortiori} arguments. The model uses existing annotations and generated explanations to analyze sentences. A new sentence is then crafted based on a chosen strategy and analyzed in a similar fashion. The results are compiled in a new dataset, compatible with the original. This process addresses the limitations discussed in Section~\ref{sec:dataset}, utilizing strategies like Semantic Similar Sentence Creation and Novel Sentence Creation to diversify sentence types and topics. The prompt's modular design facilitates the easy adaptation of the existing explanatory prompt to serve this augmentation purpose.}
    \label{fig:aug}
\end{figure}

Leveraging the modular architecture of the original prompt, the data augmentation prompt was integrated into the established framework for generating explanations. This integration involved introducing new sections and strategically reorganizing certain existing modules. Given the prior optimization of the explanation-generating prompt, the augmented version effectively produced the desired outputs without necessitating further fine-tuning.

\section{Evaluation Framework}

\subsection{Assessment of Reasoning Steps and Generated 
 Explanations}\label{sec:eva_explanation}

 Designing an  objective evaluation framework for \textit{a fortiori} logic and its explanations is challenging due to its multifaceted nature, diverse manifestations, domain-specific and general knowledge involved in the interpretation process, and the inherent subjectivity of individual annotators. The prompt learning approach, which departs from the traditional pretraining-finetuning model, only heightens these challenges. In this methodology, the model's reasoning leans heavily on its pretrained knowledge. The lack of backpropagation or parameter modifications denotes a dearth of explicit oversight of the model's actions, further complicating the process. Coupled with the lack of standardised evaluation metrics or automated tools, assessing the accuracy and quality of explanations becomes demanding.

To navigate these challenges, I introduce a hybrid evaluation approach inspired by \citet{Receval2023}. This framework integrates both automatic and human evaluations to evaluate the correctness of intermediate reasoning steps and the quality of resultant explanations. The objective is to thoroughly assess the GPT-3.5 model's ability to handle this intricate reasoning task. As shown in Figure~\ref{fig:explain}, some evaluation aspects can be automated by comparing model outputs with the dataset's ground truths. These aspects encompass the identification of the \textit{a fortiori} argument in sentences, the accuracy of extracting correlates and remnants and verifying sentence logic flow. Conversely, elements like hidden properties and short/long explanations primarily rely on human evaluation. For hidden properties, the emphasis is on relevance and contextual appropriateness, while for explanations, the assessment focuses on grammatical and logical correctness, explanation completeness, and the model's incorporation of contextual information and world knowledge for clarity and persuasiveness.

The overarching evaluation framework comprises three main components, each with multiple evaluation criteria:

\textbf{1. Evaluation of Intermediate Reasoning Steps (Automatic)}:

\begin{itemize}
    \item \textbf{Identification of \textit{a fortiori} argument}: This criterion measures the model's ability to identify \textit{a fortiori} arguments within input sentences. A confusion matrix is employed as an essential reference tool due to the presence of skewed class distributions.
    \item \textbf{Extraction Accuracy for \textit{Correlates} and \textit{Remnants}}: This evaluates how well the model can extract correlates and remnants from \textit{a fortiori} arguments. The main metrics used are semantic similarity between extracted and human-annotated items and the percentage of exact word matches.

    \item \textbf{Identification of the Sentence Type and of the Logic Category}: 
    General explanations and 10 specific examples are provided for context. A confusion matrix, combined with accuracy, precision, recall, and the F1 score, gives a full picture of the model's performance.
\end{itemize}

\textbf{2. Evaluation of Hidden Properties (Manual)}:

\begin{itemize}
    \item \textbf{Property Novelty}: This criterion assesses whether the model generates implicit comparison properties to explain the likelihood of the correlate and the remnant in cases where the property is not provided explicitly\footnote{Due to the nature of {em a fortiori} arguments; arguments with explicit properties are extremely rare}.
    \item \textbf{Property Relevance}: This evaluates whether the generated properties are relevant to the given argument.
\end{itemize}

\textbf{3. Evaluation of Short and Long Explanations (Automatic and Manual)}\footnote{The naming convention for the criteria adheres to the guidelines set out in the Validity and Novelty Prediction Shared Task \cite{heinisch-etal-2022-overview} on the assessment of the quality of implicit premises, which is a task highly relevant for my objective}:

\begin{itemize}
    \item \textbf{Syntactic Accuracy}: This automatic criterion checks the grammar and syntax of generated sentences. The LanguageTool \cite{LanguageTool} is used to spot grammatical errors. 
    \item \textbf{Logical Validity}: This evaluates if the explanation maintains inherent logical flow, specifically if it comprehends and upholds cases with varying degrees of likelihood.
    \item \textbf{Completeness}: It assesses if all crucial components (correlate, remnant, hidden properties, and relevant contexts) essential for standalone argument understanding are included in the explanation.
    \item \textbf{Contextual Information \& World Knowledge Pertinence}: This evaluates the suitability and relevance of the context and world knowledge to the argument and prevents the inclusion of irrelevant context or new knowledge.
\end{itemize}

For the evaluation, automatic methods are used on the whole dataset. However, due to time and resource constraints, manual evaluations are limited to the test set (with 100 sentences) described in Section~\ref{sec:dataset}. These evaluations are done by my principal supervisor, co-supervisor, and me. It is important to highlight that:

\begin{itemize}
    \item The criteria apply to both short and long explanations. However, in this study, only short explanations are manually evaluated. Evaluating long explanations is suggested for future work.
    \item The final point of explanation evaluation assesses the supplementary details for understanding the \textit{a fortiori} logic. This criterion can be further broken down into three distinct segments: 1) Inclusion of relevant context; 2) Offering essential world knowledge; 3) Ensuring added information isn't extraneous. If any facet is lacking in this research, the criterion is marked as "False" or “No”. Due to brevity, most short explanations focus on context over world knowledge.
\end{itemize}

\subsection{Assessment of Augmented Sentences} \label{sec:aug_evaluation_explain}

While the process of data augmentation may be more complex than explanation generation, the evaluation procedure is intentionally streamlined. This approach concentrates on aspects such as grammatical correctness, logical validity, semantic similarity between the original and augmented sentences, and the diversity and coverage of the dataset. This last aspect is evaluated by comparing the topics in the original and new sentences. Sentence topics are generated by GPT-3.5, reflecting the model's understanding, and are typically limited to one or two words. These topics are then normalized to more general terms aligned with common themes and concepts, such as `Technology,' `Education,' `Science,' and `Government \& Politics,' to facilitate further analysis.

While the augmented dataset also contains all related parts of the newly generated arguments, including their explanations, they are excluded from the evaluation at this point due to the following reasons: 

\begin{enumerate}
    \item The objective of data augmentation is to verify the model's capacity to draw parallel \textit{a fortiori} arguments from the given example. The model's reasoning capability on \textit{a fortiori} logic has been extensively evaluated in Section~\ref{sec:eva_explanation}.
    \item The analysis for the new sentences would inevitably contain errors and noise. Due to the inherent nature of prompt-based learning, it's impossible to enforce model error correction. Meanwhile, manual error correction would be excessively time-consuming and expensive.
    \item Despite the potential noise in the augmented dataset, it does not preclude its use for training or fine-tuning purposes. The maturing field of weakly supervised learning allows models to robustly solve downstream tasks from incomplete, noisy, and imprecise supervision signals in real-world scenarios, where perfectly labelled data is difficult or expensive to obtain, as in this case.
\end{enumerate}

To evaluate the grammatical correctness of the new sentences and the semantic similarity between the original and new sentences, I applied the same methods that are used for evaluating the interpretation performance (Section~\ref{sec:eva_explanation}). For assessing logical validity, 50 random sentences created by each augmentation strategy were selected for human evaluation. The primary goal here was to verify whether the model followed the augmentation instructions and whether the scenario with a higher likelihood in the sentence truly corresponds to real-world probability.

\chapter{Experiment Results And Discussions}
As discussed in Section~\ref{sec:prompt_design}, two distinct experimental approaches are used to generate explanations for \textit{a fortiori} arguments. In the first approach, no additional information is provided except for the original sentence, and the model is prompted to reason through the given sentence based on the reasoning chain supplied within the prompt, generating outputs at each step. In the second approach, all available external information from the dataset is incorporated into the reasoning process to guide the model in correctly reasoning and interpreting the given argument. Under this setting, the model is tasked solely with generating the explanations, assuming it can effectively follow the guidance and incorporate the provided external information.

\section{Model Performance on Intermediate Reasoning Steps in \textit{a fortiori} Arguments}

\begin{table}[ht]
\centering
\begin{minipage}{0.45\textwidth}
\centering
\begin{tabular}{c|ccc|c}
 & \multicolumn{3}{c|}{Ground Truth} & Total  \\ 
\hline
Prediction            & AF  & NAF & Unknown               & Total                 \\ 
\hline
AF                    & 50  & 7   & 0                     & 57                    \\
NAF                   & 660 & 40  & 0                     & 700                   \\
Unknown               & 255 & 18  & 0                     & 273                   \\ 
\hline
Total                 & 965 & 65  & 0                     & 1030   \\              \hline
\end{tabular}
\caption{The confusion matrix for identifying \textit{a fortiori} arguments in the dataset without any examples provided for guidance. Temperature = 0.3.}
\end{minipage}
\hfill
\begin{minipage}{0.45\textwidth}
\centering
\begin{tabular}{c|ccc|c} \label{tab:low_temp_withexample}
 & \multicolumn{3}{c|}{Ground Truth} & Total  \\ 
\hline
Prediction            & AF  & NAF & Unknown               & Total                 \\ 
\hline
AF                    & 483 & 31  & 0                     & 514                   \\
NAF                   & 308 & 25  & 0                     & 333                   \\
Unknown               & 175 & 8   & 0                     & 183                   \\ 
\hline
Total                 & 966 & 64  & 0                     & 1030    \\            
\hline
\end{tabular}
\caption{The confusion matrix for identifying \textit{a fortiori} arguments in the dataset with 10 examples provided for guidance. Temperature = 0.3.}
\end{minipage}
\end{table}

\subsection{Identification of \textit{a fortiori} Argument}

\begin{table}[ht] \label{tab:high_temp_identify}
\centering
\begin{minipage}{0.45\textwidth}
\centering
\begin{tabular}{c|ccc|c} \label{tab:low_temp_identify}
 & \multicolumn{3}{c|}{Ground Truth} & Total  \\ 
\hline
Prediction            & AF  & NAF & Unknown               & Total                 \\ 
\hline
AF                    & 50  & 5   & 0                     & 55                    \\
NAF                   & 877 & 54  & 0                     & 931                   \\
Unknown               & 39  & 5   & 0                     & 44                    \\ 
\hline
Total                 & 966 & 64  & 0                     & 1030   \\              \hline
\end{tabular}
\caption{The confusion matrix for identifying \textit{a fortiori} arguments in the dataset without any examples provided for guidance. Temperature = 1.}
\end{minipage}
\hfill
\begin{minipage}{0.45\textwidth}
\centering
\begin{tabular}{c|ccc|c}
 & \multicolumn{3}{c|}{Ground Truth} & Total  \\ 
\hline
Prediction            & AF  & NAF & Unknown               & Total                 \\ 
\hline
AF                    & 477 & 33  & 0                     & 510                   \\
NAF                   & 387 & 23  & 0                     & 410                   \\
Unknown               & 102 & 8   & 0                     & 110                   \\ 
\hline
Total                 & 966 & 64  & 0                     & 1030    \\            
\hline
\end{tabular}
\caption{The confusion matrix for identifying \textit{a fortiori} arguments in the dataset with 10 examples provided for guidance. Temperature = 1.}
\end{minipage}
\end{table}

\begin{table}
\centering
\begin{tabular}{l|ll|ll} 
\hline
                & \multicolumn{2}{l|}{Classification without Examples} & \multicolumn{2}{l}{Classification with Examples}  \\ 
\hline
                & AF     & NAF                                         & AF     & NAF                                      \\ 
\hline
Macro\_Accuracy & 0.0874 & 0.0874                                      & 0.4932 & 0.4932                                   \\
Precision       & 0.8772 & 0.0571                                      & 0.9396 & 0.0751                                   \\
Recall          & 0.0705 & 0.6154                                      & 0.6102 & 0.4464                                   \\
F1-Score        & 0.1304 & 0.1034                                      & 0.7388 & 0.1315                                   \\
\hline
\end{tabular}
\caption{Statistical analysis for identifying \textit{a fortiori} arguments under different experiment settings at Temperature = 0.3}
\end{table}

Although GPT-3.5-turbo has shown a satisfactory understanding of \textit{a fortiori} arguments and demonstrated capability in reasoning such arguments, as detailed in Appendix~\ref{appendix:understand_afortiori}, the initial experimental results (Table~\ref{tab:low_temp_identify}) yield unexpected findings. The model appears predominantly biased towards predicting most instances as NAF (\textit{non a fortiori}), resulting in a high number of false positives and negatives. This pattern diverges from what is typically observed with traditional training on an extremely imbalanced dataset, where an AI system might tend to assign the most common label to all unseen instances. This unusual behaviour may be attributed to the unique mechanism employed by GPT-3.5-turbo, wherein it processes each instance independently, devoid of any contextual knowledge or memory about other instances.  The tendency towards biased predictions persisted across various experiments, regardless of temperature variations, as substantiated by the results in Table~\ref{tab:high_temp_identify}.

Concurrently, the model often mislabels certain sentences as “Unknown” \footnote{Instances where the model responded with “Based on the provided information, it is not possible to determine whether the sentence contains an \textit{a fortiori} argument” were classified as “Unknown.”}, an unexpected behaviour given the binary classification required at this stage. This inconsistency is evident even in the labelling of relatively simple and intuitive sentences (e.g., example~\ref{exam:simple_sent}), where the model's reasoning appears sound with more complex arguments.
\begin{example} \label{exam:simple_sent}
    It's hard enough coming here with 11 men, let alone playing them with just ten.
\end{example}

To address the inconsistencies observed in classification, I incorporated 10 examples into the prompt to guide the model's classification -- seven of these contained \textit{a fortiori} arguments, while the remaining three did not. These examples were randomized to minimize potential false intuition to the model. As evidenced in Table~\ref{tab:low_temp_withexample}, introducing these few-shot demonstrations led to a remarkable improvement in the model's performance. The overall prediction accuracy escalated from 8.74\% to nearly 50\%, and the F1 score for AF increased significantly from 13.04\% to 73.55\%, attributable mainly to the marked increase in recall. Despite these advancements, the model continues to face challenges in classifying non \textit{a fortiori} arguments, often misclassifying them as either NAF or Unknown. This propensity to label more sentences as “Unknown” under the low-temperature setting might reflect the model’s conservative and “safe” decision-making process. Considering the unpredictability of the predictions and the limited number of experiments \footnote{OpenAI charges for every token consumed by any GPT model. Multiple runs of experiments are not affordable for an individual student who has no funding for this research project.}, it remains inconclusive whether the model's responses are based on random guesses where examples have been provided.


\subsection{Extraction of Correlates \& Remnants}

In view of the system’s inconsistent recognition of \textit{a fortiori} arguments, I tailored the guiding prompt. This modification motivated the model to evaluate each sentence, irrespective of its inclusion of an \textit{a fortiori} argument. Subsequently, \textit{non a fortiori} arguments were excluded from the study, retaining 966 instances. The assessment of the refined correlates and remnants relied on their semantic similarity which is determined by their embeddings, and the percentage of exact word matches with human annotations. These embeddings were produced using the \textit{'all-mpnet-base-v2'} model from Sentence-Transformer \cite{reimers2019sentencebert}.

\begin{table}[ht]
\centering
\begin{tabularx}{\textwidth}{XXXXX}
\toprule
\textbf{Statistic} & \textbf{Correlate Similarity} & \textbf{Remnant Similarity} & \textbf{Correlate Exact Matches} & \textbf{Remnant Exact Matches}  \\
\midrule
Mean               & 0.7163                        & 0.8414                      & 0.7049                           & 0.8321                          \\

Median             & 0.8366                        & 1.0000                      & 0.8571                           & 1.0000                          \\

Standard Deviation & 0.3177                        & 0.2664                      & 0.3432                           & 0.2920                         \\
\bottomrule
\end{tabularx}
\caption{Statistical Evaluation of Correlate and Remnant Extraction.}
\label{tab:semantic_sim}
\end{table}

Overall, the model exhibited satisfactory performance in extracting correlates and remnants, as evidenced by the strong alignment with the human-annotated data, see Appendix~\ref{Appendix:stat_analysis} for complete results of the statistical analysis. Notably, the majority of the remnants were found adjacent to the \textit{a fortiori} indicator `let alone'. This proximity could explain the model's superior performance in remnant extraction compared to correlates. In certain instances, despite a stark deviation of the extracted correlates from the ground truth, the remnants were captured flawlessly. This emphasises the pivotal role of the term `let alone' in the extraction process. Out of the total, 451 entries perfectly matched human-annotated correlates, and 651 matched remnants. On the contrary, there were instances (82 for correlates and 46 for remnants) where the model completely failed. Variability in the standard deviation primarily arises from these failure cases. 

Upon examining the failure instances, there were no identifiable patterns or links to specific sentence types or logical categories. Two primary issues surfaced: 1) The model's inability to understand certain sentences, especially those that were lengthy or complex (See Example~\ref{exam:complex}). 2) Incorrect ordering of extracted correlates and remnants (See Example~\ref{exam:reversed}). A possible reason for the latter issue might be the model's incorrect assumption that remnants refer to the more probable scenario in a comparison. Such errors, however, were infrequent and exhibited no correlation with sentence type or logical category.

\begin{example} 
    Our research shows that Israel's claim that Hezbollah fighters are hiding among civilians does not explain, let alone justify, Israel's indiscriminate warfare.

    \centering
    \begin{tabular}{|l|p{4.5cm}|p{9cm}|}
    \hline
    & \textbf{Human Annotation} & \textbf{Model Extraction} \\
    \hline
    \textbf{Correlate} & explain & Israel's claim that Hezbollah fighters are hiding among civilians \\
    \hline
    \textbf{Remnant} & justify & Israel's indiscriminate warfare \\
    \hline
    \end{tabular}
    \label{exam:complex}
\end{example}

\begin{example} 
    Show business is usually a meritocracy, but nobody ever said it was fair, let alone honorable.

    \centering
    \begin{tabular}{|l|p{5cm}|p{5cm}|}
    \hline
    & \textbf{Human Annotation} & \textbf{Model Extraction} \\
    \hline
    \textbf{Correlate} & fair & honorable \\
    \hline
    \textbf{Remnant} & honorable & fair \\
    \hline
    \end{tabular}
    \label{exam:reversed}
\end{example}

A remarkable observation is the model's inherent capability to accurately address most of the syntactic ellipsis constructions without explicit instructions.(See Examples~\ref{exam:ellipsis2}). An ellipsis refers to the omission of one or more necessary words in a sentence, which are assumed to be understood from the context but are not explicitly expressed \cite{merchant2013diagnosing}. Intriguingly, although the model scored lower in these cases based on the established evaluation criteria, its extracted correlates and remnants were often superior to human annotations. This finding underscores GPT-3.5-turbo's unanticipated proficiency, while also pointing to potential evaluation framework shortcomings.

The model displayed an inherent ability to accurately tackle most syntactic ellipsis issues without explicit guidance (See Examples~\ref{exam:ellipsis2}). An ellipsis signifies the exclusion of one or more necessary words in a sentence, inferred from the context but not explicitly stated \cite{merchant2013diagnosing}. Interestingly, although the model garnered lower scores in these instances based on the established evaluation criteria, its extracted correlates and remnants often surpassed human annotations. Another important observation relates to the imperfections present within some annotations, as highlighted in Example~\ref{exam:imperfect}. The annotated correlate 'such a move' does not parallel the subsequent action, making the reference ambiguous. However, the model demonstrated a superior ability to interpret parallel constructions and extracted a more suitable correlate term. Despite this, the model received a lower score due to the established evaluation criteria. It is plausible to assume that this imperfection is not unique and similar issues may also be present in other parts of the annotation. Consequently, additional efforts to rectify these flaws are necessary. These observations underscore the unexpected proficiency of GPT-3.5-turbo, and simultaneously bring to light potential shortcomings within the existing dataset and proposed evaluation criteria.



\begin{example} \label{exam:ellipsis2}
    We shouldn't be doing the same look architecturally or otherwise that we did 20 years ago, let alone 10 or even five years ago, said General Growth CEO John Bucksbaum.

    \centering
    \begin{tabular}{|l|p{3cm}|p{8cm}|p{2.5cm}|}
    \hline
    & \textbf{Human Annotation} & \textbf{Model Extraction} & \textbf{Semantic Similarity}\\
    \hline
    \textbf{Correlate} & 20 years ago & doing the same look architecturally or otherwise that we did 20 years ago & 0.2584\\
    \hline
    \textbf{Remnant} & 10 or even five years ago & doing the same look architecturally or otherwise that we did 10 or even five years ago & 0.2049 \\
    \hline
    \end{tabular}
\end{example}

\begin{example} \label{exam:imperfect}
    Although the revised laws allow sending peace-keeping forces under the authorization of the United Nations, such a move is still sensitive, let alone the use of weapons by SDF members.

    \centering
    \begin{tabular}{|l|p{5.5cm}|p{8cm}|}
    \hline
    & \textbf{Human Annotation} & \textbf{Model Extraction} \\
    \hline
    \textbf{Correlate} & such a move & sending peace-keeping forces under the authorization of the United Nations \\
    \hline
    \textbf{Remnant} & the use of weapons by SDF members & the use of weapons by SDF members \\
    \hline
    \end{tabular}
\end{example}

Compared with the experiments conducted by \citet{razuvayevskaya_2022} in her thesis, where she addressed the correlate and remnant detection problem as a sequence labelling task and trained several models using the argument corpus. GPT-3.5-turbo achieved comparable performance to her best model \textit{(BiLSTM-CRF)} when evaluated using the most rigorous token-wise accuracy. However, given the differences in experiment settings and instances being evaluated, this comparison may not be entirely fair. Nevertheless, the performance exhibited by GPT-3.5-turbo is noteworthy, especially considering that it did not require training or fine-tuning.

\subsection{Sentence Type and Logic Category Classification}

\begin{table}[ht]
\centering
\begin{tabular}{lcc}
\hline
 & Overall Accuracy & Overall F1 Score \\
\hline
Sentence Class Classification & 0.6677 & 0.3370 \\
Logic Category Classification & 0.4553 & 0.2806 \\
\hline
\end{tabular}
\caption{Overall Performance of the Model for Sentence Class and Logic Category Classification.}
\label{tab:overall_sent_logic}
\end{table}

To assess whether the model can correctly comprehend the explanations for sentence type and logic category, the prompt guides it to make predictions on these two tasks. Table~\ref{tab:stat_sent_type} and Table~\ref{tab:stat_logic_type} present the statistical analysis of the model’s performance on different sentence types and logic categories and Table~\ref{tab:overall_sent_logic} shows the overall performance of the model across all types \footnote{To maintain consistency with subsequent evaluations on hidden properties and explanations, these tables display the results for the specific run of the experiment used in the human evaluation. Unlike identifying \textit{a fortiori} arguments, the model's performance for these two tasks demonstrated relatively consistent performance.}; additional experiment results can be found in Appendix~\ref{Appendix:Sent_logic}. The model's performance varies across different classes; it achieves relatively good results in detecting Resource Allocation (RE) sentences, with balanced precision and recall. It also excels in classifying Negative Simple logic, as the F1 score for this type is significantly higher than other categories. However, the model struggles with instances that do not belong to any predetermined type (labelled as “Undefined”), often misclassifying them into specific types or logic categories.

\begin{table}[ht] 
\centering
\begin{tabular}{l|cccc}
\hline
Sentence Type & Accuracy & Precision & Recall & F1 Score \\
\hline
Resource Allocation            & 0.5578   & 0.5403    & 0.3389  & 0.4166   \\
Precondition            & 0.7833   & 0.5909    & 0.1135  & 0.1905   \\
Quantity            & 0.7353   & 0.3072    & 0.6667  & 0.4206   \\
Specificity            & 0.6794   & 0.2029    & 0.5966  & 0.3028   \\
Undefined     & 0.9480   & 0.3333    & 0.0600  & 0.1017   \\
\hline
\end{tabular}
\caption{Statistical Analysis for Model Performance in Sentence Type Classification}
\label{tab:stat_sent_type}
\end{table}

\begin{table}[ht] 
\centering
\begin{tabular}{l|cccc}
\hline
Logic Category & Accuracy & Precision & Recall & F1 Score \\
\hline
Negative Simple             & 0.3534   & 0.6213    & 0.2309 & 0.3367   \\
Negative Reversed             & 0.5320   & 0.1261    & 0.3958 & 0.1913   \\
Positive Reversed             & 0.8058   & 0.0760    & 0.2364 & 0.1150   \\
Positive Simple             & 0.8621   & 0.0741    & 0.1600 & 0.1013   \\
Undefined      & 0.9320   & 0.1111    & 0.0612 & 0.0789   \\
\hline
\end{tabular}
\caption{Statistical Analysis for Model Performance in Logic Category Classification}
\label{tab:stat_logic_type}
\end{table}

Compared to random guessing, the model performs remarkably better in both tasks. However, its performance in logic classification is notably poorer than the Bert-based models fine-tuned in \citet{razuvayevskaya_2022}'s thesis, which achieved F1 scores consistently higher than 0.61 and significantly outpaced GPT-3.5-turbo. The imbalanced distribution of the dataset, along with sentence type and logic category dimensions, should not affect the model performance on these tasks, as the model processes each instance independently without retaining the memory of other instances. The unsatisfactory results may stem from an insufficient understanding of relevant concepts, suggesting that further investigation is warranted.

\section{Model Performance in Interpreting \textit{a fortiori} Arguments}

This section evaluates the model's ability to interpret \textit{a fortiori} arguments, a process that involves two interconnected tasks. The first task is to predict the hidden properties that underlie the comparison between two scenarios, and the second is to synthesize these predicted properties with previously reasoned information and general knowledge to form the final explanation. These tasks are examined through experiments conducted under two distinct settings: with or without the use of external information. The relationship between the two tasks is vital, as the quality of the explanations is heavily dependent on the accuracy of the property prediction.

\subsection{Hidden Property Prediction}

\subsubsection{Without External Information}

Out of 966 instances containing \textit{a fortiori} arguments, 88 do not include any properties identified or proposed by annotators. In this experiment, the model was urged to produce at least one property for every given sentence. Consequently, it failed to produce any properties for only six entries. The model was provided with a general explanation about hidden properties in \textit{a fortiori} arguments, along with 5 examples for a detailed demonstration. In this setting, the model produced a more diverse range of properties compared to human annotation, generating 479 distinct properties (in contrast to 208 in the existing dataset), 414 of which were previously unseen in the dataset. A comparison between the predicted properties and human-annotated ones reveals no obvious alignment, with only three entries sharing identical properties between the model's prediction and human annotation. The differences in common properties for each sentence type are detailed in Appendix~\ref{Appendix:top3property}.

A review of the generated properties reveals that the model can produce highly detailed and specific attributes such as \textit{'amount of savings and investment capital'}, \textit{'suitability for earthly life forms'}, and \textit{'severity of legal consequences'}. These findings exceed our expectations, and most of these properties fit appropriately into the given context. Furthermore, the model's generation of properties such as 'Control over media' in Example~\ref{exam:murdoch} showcases the world knowledge it has acquired, including information about Rupert Murdoch.

\begin{example} 
    It's an awful lot of power in anybody's hands, let alone Rupert Murdoch's.
    \label{exam:murdoch}
\end{example}

However, the model often conservatively introduces very general terms that can be widely applied in various situations, such as \textit{Importance}, \textit{Difficulty}, and \textit{Quantity}. While these properties align with the comparative cases or scenarios in most \textit{a fortiori} arguments, the associated explanations can sound odd and unnatural. This is because these general properties are not explicitly explainable and do not enhance the explanations, even if the reasoning logic is correct. Separately, an interesting observation is that properties appearing in few-shot demonstrations are properly learned and applied in the prediction process. This is evidenced by the abnormally high frequencies of these properties in the model's predictions (see Table~\ref{tab:top10commen_woinfo}). While this demonstrates the effectiveness of few-shot learning and the model's capability in in-context learning, it raises concerns that the model's predictions may be biased toward what it has learned through few-shot examples, potentially limiting its ability to generalize beyond this specific training data.

\subsubsection{With External Information}

In this setting, the model's task is to copy properties from annotations and apply them in generating explanations, and to make its own predictions if no identified properties exist for a given sentence. As a result, 336 distinct properties were generated, 179 of which were copied from human annotations. The model exhibits good alignment with the instructions, with 755 out of 966 entries containing at least one property that matches the annotations. Among these, 685 entries have both properties that match exactly with the annotated dataset. Furthermore, the evaluation shows a nearly perfect alignment in property frequencies across all types of sentences.

\begin{example} 
    However, he emphasized that the prevention of nuclear proliferation should in no way affect or hinder normal nuclear cooperation for peaceful use among various countries, let alone be used as an excuse for discrimination or an application of willful sanctions against developing countries.
    
    \centering
    \begin{tabular}{|p{6cm}|p{7.5cm}|}
    \hline
    \textbf{Original Property} & \textbf{New Property} \\
    \hline
     Compliance with moral rules & Compliance with international agreements \\
    \hline
    \end{tabular}
    \label{exam:nuclear}
\end{example} 

However, the model does not rigidly adhere to the guidance of copying every single property. Instead, it occasionally rejects properties from the existing dataset and proposes new ones. For instance, in Example~\ref{exam:nuclear}, the model proposed a new property, \textit{'Compliance with international agreements'}, believing it to be more suitable than the original one. Below is the model's explanation of its choice:

\begin{quote}
    `The prevention of nuclear proliferation and the promotion of peaceful nuclear cooperation are typically governed by international treaties and agreements. Adhering to these agreements ensures that actions are coordinated, transparent, and in line with globally accepted standards. While moral rules are undoubtedly important, they can be more subjective and vary between cultures and individuals. Compliance with international agreements provides a more concrete and standardized framework that can be applied uniformly across different countries.'
\end{quote}

Although the superiority of this newly generated property may be debatable, the model does generate properties that fit better than human-annotated ones (see Example~\ref{ex:better_property}) in some cases. This highlights the model's deep understanding of the task and robustness in its knowledge and decision-making process, reflecting its ability to innovate or optimize beyond human expertise. However, this capability may also raise concerns about transparency, control, and ethical considerations, especially when the underlying rationale for the AI system's decisions is not well understood, or if those decisions carry significant real-world implications.

\begin{example}
    Personally I'm delighted with the business we've done over the whole window, let alone today, including getting rid of the dead wood.
    
    \centering
    \begin{tabular}{|l|p{5cm}|p{5cm}|}
    \hline
    & \textbf{Human Annotation} & \textbf{Model Prediction} \\
    \hline
    \textbf{Property 1} & Amount of & Achievement \\
    \hline
    \textbf{Property 2} & Duration & Specificity \\
    \hline
    \end{tabular}
    \label{ex:better_property}
\end{example}

\subsubsection{Human Evaluation of Property Quality}

This part of the evaluation is based on 100 selected sentences created in Section~\ref{sec:dataset}. The focus is on a binary judgment concerning the predicted properties' novelty and relevance. General properties that can be applied in most \textit{a fortiori} arguments are labelled as relevant but not novel. Meanwhile, properties that are directly copied from human annotations are not assumed to be novel and relevant by default; they are reassessed in the given context.

\begin{figure}[ht]
    \centering
    \includegraphics[width=1\textwidth]{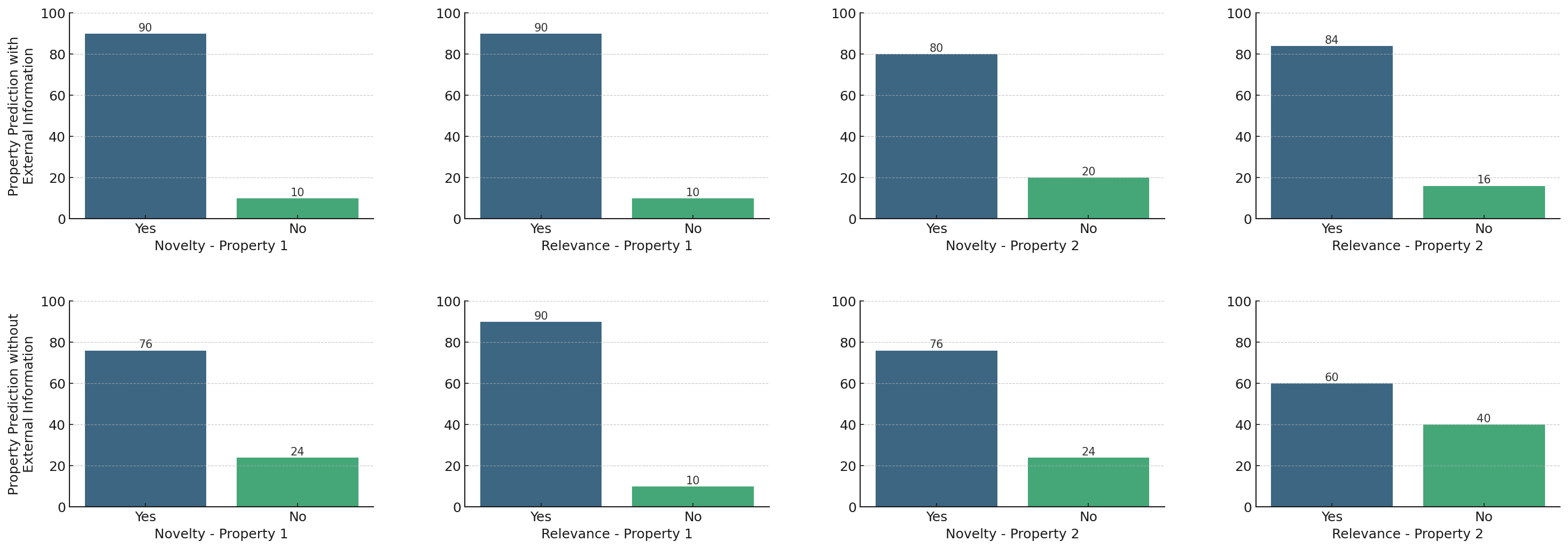}
    \caption{Model Performance in Property Quality Evaluation}
    \label{fig:property_count}
\end{figure}

Figure~\ref{fig:property_count} summarizes the results of the evaluation, and Table~\ref{tab:humaneva_property} presents a more detailed view of the evaluation results at both the sentence and property levels. Examining both the sentence and property levels is essential, as relying solely on the count of correctness could lead to a misleading interpretation of the model's performance.

\begin{table}[ht]
    \centering
    \begin{tabular}{|p{6cm}|p{3cm}|p{3cm}|p{2.5cm}|}
    \hline
    \textbf{Experiment Setting} & \textbf{With External Information} & \textbf{Without External Information} & \textbf{p-value} \\
    \hline
    Sentences with both properties novel and relevant & 54/100 & 34/100 & \(9.75 \times 10^{-5}\) \\
    \hline
    Sentences with at least one property novel and relevant & 94/100 & 80/100 & \(1.16 \times 10^{-4}\) \\
    \hline
    Sentences with both properties neither novel nor relevant & 0/100 & 2/100 & 0.1583 \\
    \hline
    Properties both novel and relevant & 94/200 & 80/200 & 0.0033 \\
    \hline
    Properties relevant but not novel & 24/200 & 32/200 & 0.0002 \\
    \hline
    Properties novel but not relevant & 20/200 & 34/200 & 0.0033 \\
    \hline
    Properties neither novel nor relevant & 4/200 & 10/200 & 0.0002 \\
    \hline
    \end{tabular}
    \caption{Human Evaluation Summary of Property Quality at both the Sentence and Property Levels. There are 200 predicted properties associated with 100 evaluated sentences. A paired t-test was conducted, with a significance level set at 0.05.}
    \label{tab:humaneva_property}
\end{table}

According to the plots, the majority of the generated properties have passed human evaluation. However, in both settings, fewer than half of the properties met the assessment criteria for both novelty and relevance. Nonetheless, further insights emerge from the statistical analysis. The paired t-test indicates that the model's performance significantly improves when external information is provided

It's worth noting that the performance in this experiment setting may not reflect the true capability of the model. Since human-annotated properties are fed into the interpretation process, and the model is allowed to directly copy them, it theoretically should perform better if it chooses to copy properties for every single sentence. In my opinion, the performance without external information might be closer to the real capability of GPT-3.5-turbo, although further investigation is needed to confirm this hypothesis. Lastly, the results of this experiment are not directly comparable to \citet{razuvayevskaya_2022}'s experiment results due to the completely different experiment design and evaluation procedure.

\subsection{Explanation Generation}\label{sec:explain_generation}

The model is configured to formulate two distinct categories of explanations. The short explanation is confined to a singular sentence, encapsulating essential elements as identified, classified, and predicted in the antecedent stages. In contrast, the long explanation affords the model an augmented latitude for argumentation, thereby allowing the integration of external knowledge beyond the immediate context for clearer elucidation. Within the purview of this research endeavour, human evaluation was conducted exclusively on short explanations for a selection of 100 instances.

The evaluation of the model explanations begins with a grammatical examination using LanguageTool \cite{LanguageTool}. As delineated in Appendix~\ref{Appendix:grammar_error}, neither short nor long explanations exhibit fundamental syntactic problems, though some spelling inconsistencies, capitalization, and whitespace issues persist, with most being inherited from the original sentences. Despite these grammatical accuracies, human evaluation has disclosed that certain explanations sound awkward or stilted, and some even lack coherence (see Example~\ref{exam:use_require} and Example~\ref{exam:use_property}).

\begin{example}\label{exam:use_require}
\textbf{Sentence}: The spokesman explained that the 1,500 families on the organization's list are all so poor that they cannot even afford a decent meal, let alone a feast.\\
\textbf{Explanation}: Affording a feast requires more cost than affording a decent meal.
\end{example}

\begin{example}\label{exam:use_property}
\textbf{Sentence}: After years of waiting for a decision, let alone action, the water-logged people of low-lying areas of Venice have had it up to here with studies and up to here with flooding.\\
\textbf{Explanation}: Taking action requires a \textbf{typical sequence of actions}, which takes more time than waiting for a decision.
\end{example}

One recurring issue is the overuse of specific terms, such as `require', in constructing explanations. This tendency can be attributed to the templates provided to the model to normalize the form of the explanation, where `require' appears multiple times across various templates and sentence types. This repetition has inadvertently led the model to favour this word in explanation formation. In addition, some explanations enforce the incorporation of identified properties, particularly general terms, without adequate paraphrasing to suit the given context. For sentences encountering issues similar to those in Example~\ref{exam:use_property}, the model often generates more meaningful explanations without the assistance of external information, as it is not constrained in the property prediction step.

The explanatory quality is evaluated on the basis of three discrete dimensions delineated in Section~\ref{sec:eva_explanation}: Logical Validity, Completeness, and the Pertinence of Contextual Information \& World Knowledge. These criteria are organized by the complexity of their corresponding requirements, with Logical Validity evaluating the basic comprehension of logic flow, and Contextual Information \& World Knowledge assessing the most intricate aspect of the explanation. For short explanations, the requirement for the last criterion was relaxed due to length constraints. If an explanation incorporates certain contextual information extracted from the given argument that aids in the understanding of the logic, it is marked as meeting this requirement. The rationale behind this approach is that it is often impractical to include all key concepts and general knowledge within a single sentence without compromising intelligibility. An explanation fulfilling Logical Validity and Completeness is deemed acceptable; if it also meets the third criterion, it is labelled as of good quality.

\begin{figure}[h!]
\centering
\includegraphics[width=1\textwidth]{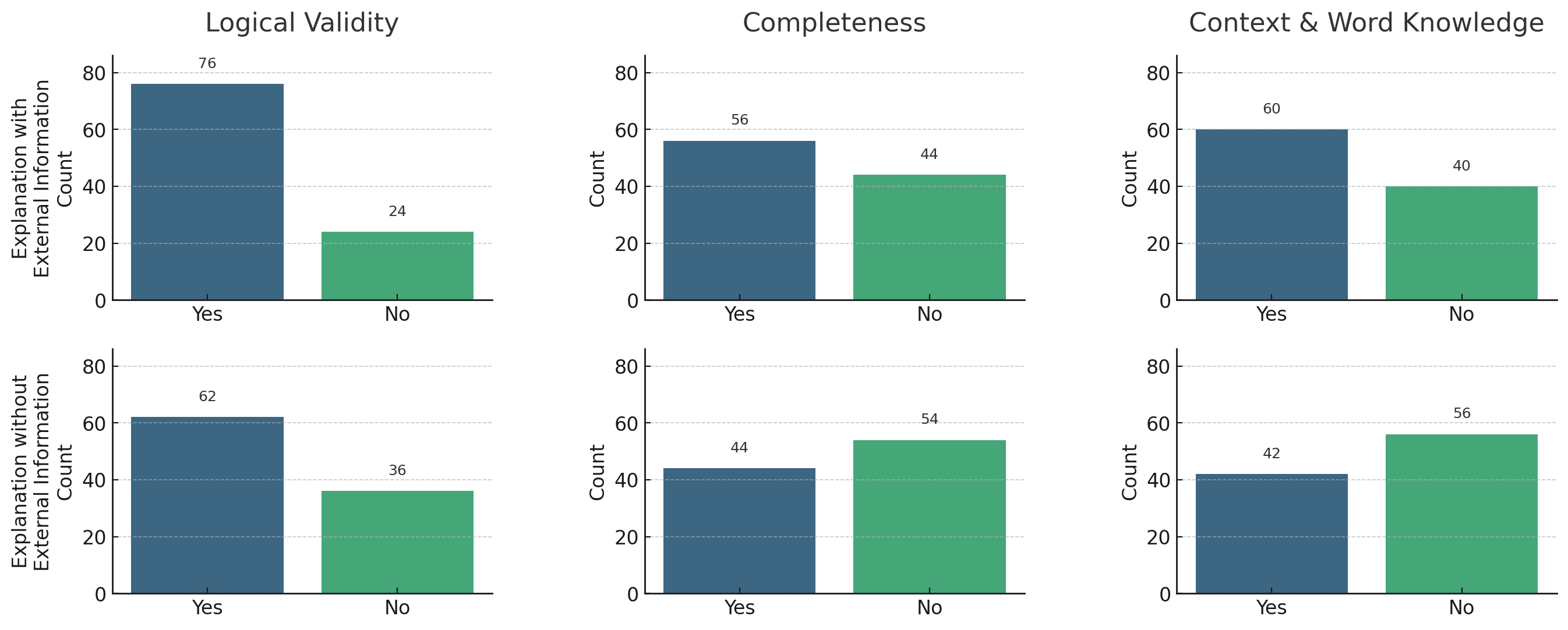}
\caption{Model Performance in Explanation Quality Evaluation}
\label{fig:explain_count}
\end{figure}

\begin{figure}[ht]
\centering
\includegraphics[width=1\textwidth]{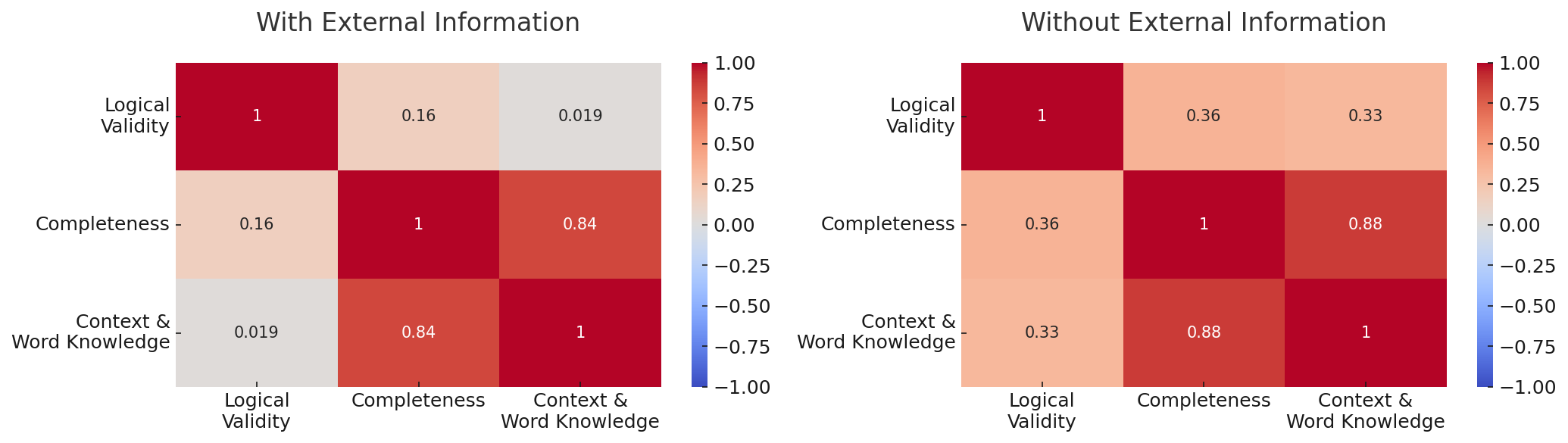}
\caption{Evaluation Criteria Correlation for Explanation Quality}
\label{fig:explain_correlation}
\end{figure}

Figure~\ref{fig:explain_count} delineates the overall performance of the model under various experimental settings. The model excels in simpler tasks, as evidenced by a higher count of affirmations for Logical Validity. However, it encounters challenges in incorporating all key concepts into the explanation, as measured by `Completeness.' Instances where the model identifies correct hidden properties for comparison but omits them in the explanation illustrate this struggle. A strong correlation between `Completeness' and `Contextual Information \& World Knowledge Pertinence,' as depicted in Figure~\ref{fig:explain_correlation}, may be attributable to our lenient assessment of the latter criteria. This unintended correlation is undesirable since the evaluation design aimed to ensure independence among the criteria, allowing for a multifaceted assessment of explanations.

Table~\ref{tab:explanation_comprehensive} offers a comprehensive overview of the evaluation results, enumerating explanations that satisfy one, two, or all three criteria. Significance tests reveal that the model's performance generally improves with the aid of external information, particularly in more complex tasks assessed by the second and third criteria. Such enhancement leads to marked progress in generating high-quality explanations, underscoring the robust correlation between hidden property prediction and explanation formulation, as well as the critical role of accurate property inference. Conversely, the integration of external information may constrain the model's behaviour to human intelligence, potentially limiting its adaptability. Remarkably, in certain instances, the model can produce superior explanations without external assistance. In Example~\ref{exam:better_explain}, the explanation generated without external information is notably more insightful than its counterpart, as it directly elucidates the relationship between stating and proving within the context of AP stories.

\begin{table}[ht]
\centering
\begin{tabular}{|p{7.5cm}|p{2cm}|p{2cm}|p{2cm}|}
\hline
& \textbf{With External Information} & \textbf{Without External Information} & \textbf{P-value} \\
\hline
Explanations passed the first criterion only & 28 & 26 & 0.1583 \\
Explanations passed both the first and second criteria but not the third & 2 & 2 & N/A \\
Explanations passed the second and third criteria but not the first & 10 & 6 & \textbf{0.0449} \\
Explanations passed the first and third criteria but not the second & 2 & 0 & 0.1583 \\
Explanations passed all three criteria & 44 & 34 & \textbf{0.0013} \\
\hline
\end{tabular}
\caption{Number of Explanations meeting one, two, or all three evaluation criteria. The first, second, and third criteria correspond to \textbf{Logical Validity}, \textbf{Completeness}, and \textbf{Contextual Information \& World Knowledge Pertinence}, respectively. A paired t-test determined significance with a level set at 0.05.}
\label{tab:explanation_comprehensive}
\end{table}

\begin{example}\label{exam:better_explain}
\textbf{Sentence}: No AP story has said Feinstein lied about this, let alone proved that she lied, said AP's San Francisco Bureau Chief Dan Day. \\
\textbf{Explanation With External Information}: Proving that Feinstein lied requires a typical sequence of actions that goes beyond simply stating that she lied in an AP story. \\
\textbf{Explanation Without External Information}: Proving that Feinstein lied requires more evidence than simply stating that no AP story has said she lied.
\end{example}

\section{Model Performance in Data Augmentation}

\subsubsection{Grammatical Correctness and Consistency}

The evaluation of augmented sentences across various strategies prioritizes their grammatical correctness, semantic alignment with original sentences, and diversification across different topics. In examining grammar, most issues identified by LanguageTool pertain to spelling inconsistencies and lowercase misuse, which do not hinder comprehension of the arguments in these sentences. It is noteworthy that similar sentences present three times as many such issues (151 errors in total) as novel sentences (49 errors in total); see Appendix .

\subsubsection{Semantic Similarity in Augmentation Strategies}

\begin{figure}[ht]
\centering
\includegraphics[width=1\textwidth]{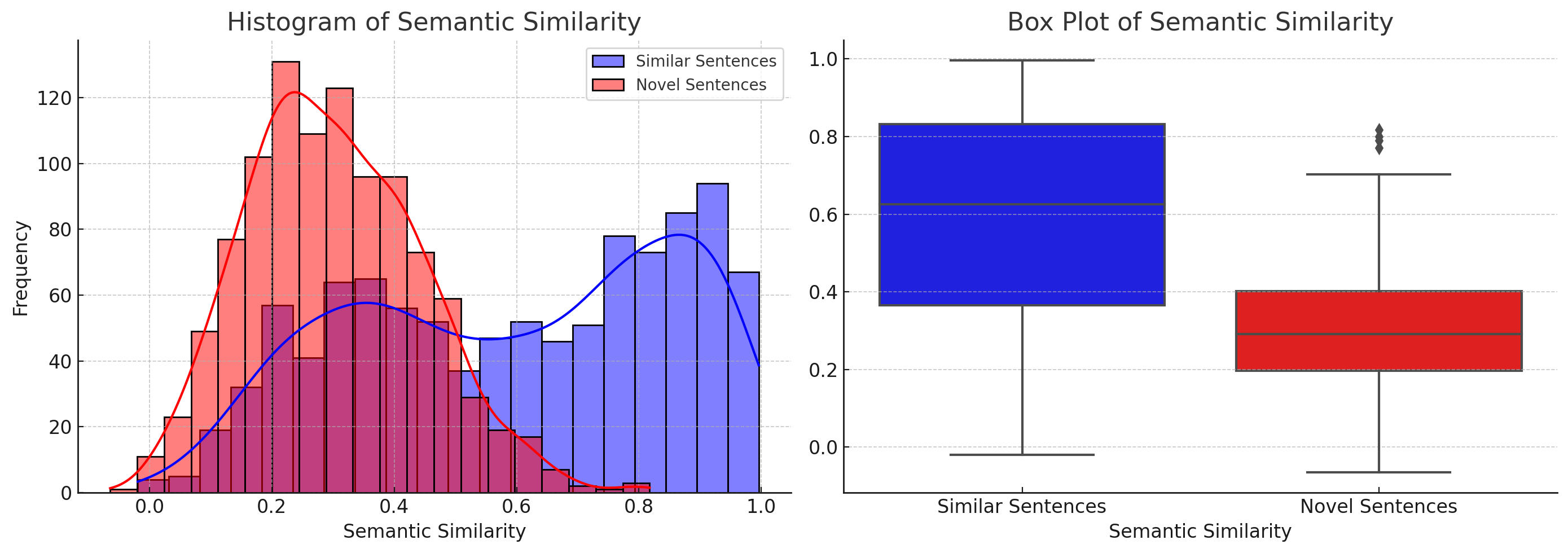}
\caption{Semantic Similarity Distribution for Augmented Sentences}
\label{fig:semantic_similarity_dist}
\end{figure}

Under the `Semantic Similar Sentence Creation' strategy, the augmented sentences moderately resemble the given ones, with a mean similarity score of 0.5956. This is considerably higher than the score for augmented novel sentences (mean similarity score = 0.3012). The distribution of scores for similar sentences is left-skewed, with a higher frequency near 1, while the distribution for novel sentences is more regular, leaning towards lower similarity scores. This pattern reaffirms the model's compliance with the guidelines for data augmentation. However, there have been instances where augmented sentences created under the similar augmentation strategy are entirely unrelated to the original sentence. This discrepancy, illustrated in Example~\ref{exam:dissimilar_sent}, warrants further examination.

\begin{example} \label{exam:dissimilar_sent}
    Original Sentence: Mario Cart will give me so much stress already let alone being on the damn road.\\
    Augmented Sentence: Playing the piano for a concert requires a lot of practice, let alone performing in front of a large audience.
\end{example}

To create semantically similar sentences, the model employs various techniques, including paraphrasing the entire sentence, paraphrasing the correlate and remnant, substituting these terms with comparable alternatives, using different comparative indicators instead of "let alone," or a combination of these techniques. Example~\ref{exam:augment_strategy}  illustrates typical methods of constructing semantically similar sentences. Consequently, the use of 'let alone' for constructing comparisons is less prevalent in this context (704 out of 966) than in the novel sentence augmentation, where all sentences adhere to this guidance.

\begin{example}\label{exam:augment_strategy}
    Original Sentence: Ambassador to the International Atomic Energy Agency (IAEA) and the United Nations in Vienna, expressed his lack of optimism that the NPT review conference would have any bearing on the Iran and DPRK nuclear issues, let alone produce a strengthened Treaty.\\
    Augmented Sentence: The ambassador to the International Atomic Energy Agency (IAEA) and the United Nations in Vienna, expressed his lack of optimism that the NPT review conference would have any impact on the Iran and DPRK nuclear issues, much less result in a comprehensive agreement.
\end{example}

\subsubsection{Topic Diversity and Logical Flow}
In the novel sentence augmentation setting, the model generates more varied sentences across different topics \footnote{The most frequent topics occurring in each augmentation setting are detailed in Table~\ref{fig:common_topic}}, with 415 out of 535 emergent topics, enhancing the original dataset’s diversity. Though normalisation preceded the analysis of topic coverage, it must be noted that similar topics may still represent the same concept.

\begin{table}[ht]
\centering
\begin{tabular}{|p{6cm}|p{3.5cm}|p{3.5cm}|}
\hline
 & \textbf{Augmented Similar Sentences} & \textbf{Augmented Novel Sentences} \\
\hline
Unique Raw Topics & 700 & 790 \\
Unique New Topics & 642 & 535 \\
Same Topics & 314 & 121 \\
Emergent Topics & 329 & 415 \\
"Let Alone" in New Sentences & 704 & 966 \\
\hline
\end{tabular}
\caption{Comparison of Topics in Augmented Similar and Novel Sentences}
\label{tab:augmented_sentences_comparison}
\end{table}

Upon randomly selecting 50 sentences from each augmentation type, the human evaluation results reveal the model's accurate understanding of the terms being compared in nearly all cases (48/50 for similar sentences and 49/50 for novel sentences). Nevertheless, occasional confusion may arise in the comparison process. This is evident in Example~\ref{exam:confusion_aug}, where the elements being compared lack a clear relationship, rendering them incomparable. This research did not evaluate the augmentation process in terms of the model's ability to analyse newly augmented sentences or produce results in the same format for intermediate reasoning steps and final explanations. The reasons for this omission are explained in Section~\ref{sec:aug_evaluation_explain}.

\begin{example}\label{exam:confusion_aug}
    Original Sentence: He could no longer read their position on the Loran let alone see the coast and the mouth of the Makaa River which remained his goal.\\
    Augmented Sentence: He could no longer understand their position on the Loran, let alone reach the summit of Mount Everest, which remained his ultimate challenge.
\end{example}

\section{Limitations}

\subsection{Inconsistencies in Task Alignment}

The model occasionally exhibits inconsistencies in aligning with the instructions in the prompts during experiments. Specific abnormalities observed include:

\begin{enumerate}
    \item Failure to produce predictions or explanations for some instances.
    \item Inconsistency in following instructions to replicate human-annotated properties in property prediction tasks.
    \item Occasional lack of application of predicted properties to explanations.
    \item Inconsistent generation of semantically similar augmented sentences as instructed.
    \item Failure to produce semantic similar sentences
\end{enumerate}

The first issue was addressed by utilizing the \textit{gpt-3.5-turbo-16k-0613} model, featuring a window size four times larger than the default model. This measure was necessary as the model would otherwise cease reading or generating content upon reaching its limit, thereby resulting in incomplete results or an inability to process the given instance.

The remaining issues persist across different experimental runs and are beyond human intervention through prompting. Two factors may contribute to these issues:
\begin{enumerate}
    \item \textbf{Weak supervision inherent to prompt learning.} Though the rich context and examples in the prompt allow the model to adapt to various tasks without specific task-oriented training, it has not been explicitly trained on these tasks, causing failures to follow instructions precisely.
    \item \textbf{Sensitivity demonstrated in empirical experiments \cite{chen2023robust}}. Although GPT-3.5 shows robust performance on numerous NLI tasks, it is sensitive to task instructions, prompt strategies, and data or label types. This sensitivity may contribute to the observed inconsistencies. The underlying mechanisms, however, remain unclear and call for further exploration.
\end{enumerate}

\subsection{Shortcomings in Prompt Design}

The flexibility of the modular design in this research project, even with iterative tuning, fails to fully address the prompt design deficiencies that may affect model performance. The observed deficiencies include:

\begin{itemize}
    \item \textbf{Lengthy Prompts}: To guide the model towards generating desirable outputs, complex and detailed prompts were employed, some exceeding 2500 words. The extensive length of these instructions risks overwhelming the model, causing confusion or misinterpretation, and potentially leading to truncation of information due to token limits, thereby compromising comprehension.
    \item \textbf{Complex and Interrelated Steps}: The prompts necessitate the model to process multiple complicated tasks sequentially, involving intricate and interconnected steps. Misunderstandings or improper executions within this sequence may propagate errors to subsequent stages. 
    \item \textbf{Compromised Prompt Tuning}: Despite repeated tuning of the prompts, emphasis was placed solely on conforming to correct output formats, with no assessment of prompt effectiveness or output quality. Initial intentions to scrutinise various prompting strategies were set aside because of time and budget constraints. Additionally, the complexity of the task and the absence of defined evaluation criteria render automatic tuning impracticable.
\end{itemize}

\subsection{Model Randomness in Generation and Challenges in Evaluation}

\subsubsection{Model Randomness in Generation}
As delineated in the OpenAI API reference \footnote{https://platform.openai.com/docs/api-reference}, GPT-3.5 retains a degree of randomness, even with the temperature set to zero, to enhance diversity and creativity. This inherent randomness results in varied outputs across experimental runs, impacting the accuracy of predictions and the quality of generated content. Such unpredictability complicates the fair assessment of the model's reasoning and interpretation, especially considering the labour-intensive nature of human evaluation. Furthermore, the financial burden of multiple experimental runs, stemming from token consumption, restricts individual students. Consequently, evaluations are typically based on a single run, potentially yielding one-sided and incomplete findings that may not faithfully represent the model's capabilities.

\subsubsection{Challenges in Evaluation}

Evaluating the quality of predicted hidden properties and generated explanations leans heavily on human judgment, introducing complexity and subjectivity. The acceptance of multiple interpretations for a single \textit{a fortiori} argument adds to the variability. Complications arose from unforeseen reductions in the number of annotators and the lack of calculation for inter-annotator agreement. Even though a consensus exists among annotators, unresolved disagreements in judgment further muddy the evaluation process.

\subsubsection{Evaluation Framework Issues}

The evaluation framework proposed, with the intention of establishing a universal standard for quality assessment, exhibits inherent shortcomings. Section~\ref{sec:explain_generation} emphasizes a significant correlation between \textit{Completeness} and \textit{Contextual Information \& World Knowledge Pertinence}, revealing overlaps that conflict with the intended differentiation. The necessary improvements and refinements to this framework must be supported by further evidence.

\subsection{Failure in Fine-tuning Text-to-Text Models}

The initial proposal set forth an ambitious objective: to fine-tune smaller text-to-text models like T5 or FLAN-T5. The goal was to generate explanations for \textit{a fortiori} arguments, using the guidance of explanations previously generated by GPT-3.5. Extensive experiments were conducted across a range of T5 and FLAN-T5 models of different sizes, from the smallest model T5-small (with 60.5M parameters) to the largest model accessible with the available computational resources — FLAN-5-XXL (with 11.3B parameters). All augmented instances were included during the fine-tuning process to provide additional examples. Unfortunately, none of the models yielded meaningful explanations for the provided sentences. Instead, they merely paraphrased the sentences, failing to improve the understanding of the arguments. The failure can be attributed to several factors:

\begin{itemize}
    \item \textbf{Model Scale and Pretraining Tasks}: Past experiments \cite{huang2023reasoning} indicate that competent reasoning abilities typically manifest in language models exceeding 100 billion parameters. Models below this threshold often failed to generate the desired explanations. Both T5 and FLAN-T5 were pretrained on diverse but relatively simplistic tasks, which did not equip them with the nuanced understanding required for our complex reasoning task. Neither demonstrated an innate grasp of \textit{a fortiori} logic or an ability to create sophisticated explanations.
    \item \textbf{Dataset Size and Model Structure}: Our dataset, containing just a few thousand samples across 25 combinations (5 sentence types * 5 logic types), paled in comparison to other successful fine-tuning cases for reasoning tasks. This may have left the models unequipped to comprehend the complexity of the assignment. Further compounding the problem was the internal structure of the T5 variants. Their encoder-decoder architecture did not facilitate step-by-step guidance through the reasoning process, unlike GPT-3.5 and GPT-4, which can be incrementally led to its conclusions.
\end{itemize}

These underlying factors elucidate the challenges faced during the fine-tuning process. While the attempt did not yield the anticipated results, the insights gained may serve to guide future endeavours in leveraging smaller text-to-text models for complex reasoning and explanation generation tasks.

\chapter{Conclusion and Future Work}

This research embarked on an exploration into the complex domain of \textit{a fortiori} arguments, with the aim to utilize the GPT-3.5-turbo model to understand and interpret \textit{a fortiori} arguments. The significance of this thesis is manifested in its academic novelty and its potential applications across diverse fields. By translating complex human thought into a form that large language models (LLMs) can handle through various prompting strategies, this research contributes to the broader understanding of logical reasoning and its automation.

This thesis implemented a series of meticulously designed experiments to evaluate the model's performance in the intermediate reasoning steps, interpretation, and augmentation of \textit{a fortiori} arguments. These comprehensive experiments, covering various aspects like argument identification, correlate \& remnant extraction, argument classification, hidden property prediction, explanation generation, and data augmentation, reflect the thesis's determination to offer a broad and detailed view of the model's capabilities and limitations. While highlighting successes in areas such as extracting correlate \& remnant and predicting hidden properties, it also illuminated shortcomings, such as misclassifications and unnatural phrasing. Moreover, this research tackled data scarcity in logical reasoning research by utilizing data augmentation to enhance the diversity of the existing dataset.

The explorative nature of this research leaves several promising avenues for future work that may enhance the model's capability in automating reasoning and interpreting \textit{a fortiori} arguments:

\begin{itemize}
    \item \textbf{Exploration of Advanced Models}: GPT-4 has demonstrated significant improvements in comprehension, contextual understanding, and logical reasoning. Re-conducting experiments using these models could lead to notable enhancements in both accuracy and efficiency. The resulting insights may uncover unseen aspects of \textit{a fortiori} arguments, providing more robust and nuanced explanations. By keeping pace with the latest advancements in the field, this research direction could reinforce the foundations of automated logical interpretation, contributing to the ongoing evolution of AI-driven logical reasoning.
    \item \textbf{Dataset Enrichment and Diversification}: A rich and diverse dataset is vital for a comprehensive understanding of complex logical constructs. By including more training instances and additional layers of complexity, such as short and long explanations and sentence topics, the model's comprehension and interpretation capabilities could be significantly sharpened. This enrichment not only lays a more solid groundwork for future advancements but also widens the scope of applicability, addressing the data scarcity issue and enhancing the diversity of the existing dataset.
    \item \textbf{Objective and Automated Evaluation Methods}: Current evaluations that rely heavily on human judgments can introduce inconsistencies, hindering the replicability and scalability of the research. Developing automatic, objective, and consistent evaluation techniques with minimal human intervention would provide a more transparent and universally applicable assessment of the model's reasoning abilities. Such refinement could lead to broader acceptance and application of \textit{a fortiori} argument analysis in various domains.
    \item \textbf{Interactive \& Automatic Prompt Optimization}: The reasoning behind \textit{a fortiori} arguments is complex, requiring a nuanced approach. By breaking down the reasoning process into interconnected, evaluated steps, automatic prompt fine-tuning becomes possible. This structured approach could enhance the model's adaptability and precision, optimizing guidance at each stage, and mitigating the risk of cascading mistakes. Implementing this multi-step method could create a more agile and reliable reasoning engine, representing a promising avenue for research and development.
    \item \textbf{Adoption of Knowledge Distillation Techniques}: The challenges associated with fine-tuning large models and deploying them in constrained environments may be addressed through Knowledge Distillation. This process of transferring specific reasoning knowledge from large models to smaller ones offers a pathway to maintain performance without the constraints of size \cite{gou2021knowledge}. It can lead to faster convergence, better generalization, and adaptability across various platforms \cite{ho2023reasoning_teacher}. Exploring Knowledge Distillation in the context of \textit{a fortiori} arguments could unlock new potentials, making logical reasoning tasks more accessible and efficient.
\end{itemize}

In summation, this thesis represents a seminal contribution to the intersection of logic, language, and artificial intelligence. By investigating \textit{a fortiori} arguments, the study reveals both opportunities and challenges in modelling the intricate human reasoning process. Far beyond mere academic endeavour, this research is a pioneering effort with potentially transformative impacts across various domains. The questions raised and the new paths opened signify a promising beginning, heralding new horizons in automated logical reasoning.

\label{lastcontentpage} 

\begingroup
\raggedright
\bibliographystyle{unsrtnat}
\bibliography{reference}
\addcontentsline{toc}{chapter}{Bibliography} 
\endgroup

\appendix
\addcontentsline{toc}{chapter}{Appendix}

\chapter{Components in Prompt Design}

\section{Preliminary Experimental Results: Performance of GPT Variants in Explaining \textit{a fortiori} Arguments at Varying Temperatures} \label{Appendix:preli_test}
\begin{table}[ht]
    \centering
    \begin{tabular}{llllllllllllll}
    \toprule
    & & \multicolumn{2}{c}{GPT3.5–Default} & \multicolumn{2}{c}{GPT3.5–Low Temp.} & \multicolumn{2}{c}{GPT4–Default} \\
    \cmidrule(lr){3-4} \cmidrule(lr){5-6} \cmidrule(lr){7-8}
    Category & Sent No. & Short & Long & Short & Long & Short & Long \\
    \midrule
    SP & 1 & 0 & 1 & -1 & -1 & 0 & 2 \\
     & 2 & 0 & 0 & 1 & 2 & 1 & 1 \\
     & 3 & 0 & 0 & 0 & 1 & 0 & 2 \\
    RE & 4 & 0 & 0 & 0 & 1 & 0 & 2 \\
     & 5 & 0 & 1 & 0 & 2 & 0 & 2 \\
     & 6 & 0 & 1 & 0 & 2 & 0 & 2 \\
    PC & 7 & 0 & 1 & 0 & 1 & 0 & 1 \\
     & 8 & 0 & 1 & 0 & 1 & 0 & 2 \\
     & 9 & 0 & 2 & 0 & 2 & 0 & 2 \\
    QU & 10 & 0 & 1 & 0 & 1 & 0 & 2 \\
     & 11 & 0 & 1 & 0 & 2 & 0 & 2 \\
     & 12 & 0 & 1 & 0 & 1 & 0 & 2 \\
     \midrule
    Avg. & & 0.00 & 0.83 & 0.00 & 1.25 & 0.08 & 1.83 \\
    \bottomrule
    \end{tabular}
    \caption{Preliminary comparison of the quality of short and long explanations produced by GPT-3.5 and GPT-4 models under default and low-temperature settings. In the default setting, the temperature for both GPT-3.5 and GPT-4 is set to 1. For the low-temperature setting, the temperature is reduced to 0.1. Evaluation scores are as follows: \textbf{-1} for incorrect explanations, \textbf{0} for mere paraphrasing without added insight, \textbf{1} for acceptable but improvable explanations, and \textbf{2} for exceptional explanations that encapsulate the core concept of the argument.}
    \label{tab:pre_test}
\end{table}

\section{Templates Utilized for Normalizing Short Explanations Across Sentence Types} \label{Appendix:norm_temp}

\paragraph{QU (Quantitative)}
\begin{enumerate}
\item X is [greater/more substantial/bigger/higher] than Y in terms of [measurement P].
\item More [measurement P] is required to be X than to be Y.
\item X [is/offers/contains/provides] more [measurement P] than Y.
\end{enumerate}
Example: \\
Sentence: The armchair cannot be put into a 14 sq ft garden shed, let alone a 5 sq ft one.\\
Short Explanation: A 14 sq ft garden shed provides more space than a 5 sq ft shed.

\paragraph{RE (Resource Allocation)}
\begin{enumerate}
\item A greater amount of P is needed for X than for Y.
\item More P is required to achieve X than to achieve Y.
\item X requires more P than Y.
\item X encompasses/contains more P than Y.
\end{enumerate}
Example: \\
Sentence: The men on trial are all accused of being accessories to the murder, but the authorities have still failed to find the triggerman, let alone identify the mastermind of the killing.\\
Short Explanation: Identifying the mastermind of the killing requires more information than finding the triggerman.

\paragraph{SP (Specificity)}
\begin{enumerate}
\item X [is more/ has more/ takes more] [specific property P] than the average Y. 
\item X is more specific than Y.
\end{enumerate}
Example: \\
Sentence: Yeah, this isn't much of a deterrent for any NBA organization, let alone the Lakers...\\
Short Explanation: The Lakers possess more authority than the average NBA organization.

\paragraph{PC (Precondition)}
\begin{enumerate}
\item X must occur before Y can be achieved/done.
\item Unless X [occurs/is done], Y cannot [take place/ be done/ exist].
\item X is a prerequisite/precondition for Y.
\end{enumerate}
Example: \\
Sentence: I have never heard of it, let alone using it, she said.\\
Short Explanation: Unless one heard of something, one cannot use it.

\section{Commonly Observed Implicit/Hidden Properties in \textit{a fortiori} Arguments as Identified by \citet{razuvayevskaya_2022}} \label{Appendix:common_properties}

\begin{enumerate}
\item Number of/amount of
\item Importance
\item Priority/pyramid of needs
\item Specific case of
\item Attractiveness/appealingness
\item Financial cost
\item Financial profit
\item Typical sequence of actions
\item Effort emotional
\item Effort physical
\item Effort intellectual
\item Damage physical
\item Damage emotional
\item Familiarity/information accessibility
\item Time required
\item Skills needed
\item Proximity
\item Distance
\item Danger
\item Size
\item Other scalars (weight, temperature, volume)
\item Control level
\item Criminality/illegality
\end{enumerate}

\section{Manually-Crafted Examples Utilized in the Few-Shot Setting} \label{Appendix: examples_fewshot}

\noindent \textbf{Example1:} \\
\textbf{Sentence:} Under the former Iraqi nationality law, Iraqis are prohibited from holding dual nationality and those having dual nationality are not allowed to act as public officials, let alone legislators. \\
\textbf{Correlate:} public officials \\
\textbf{Remnant:} legislators \\
\textbf{Likelihood:} Yes \\
\textbf{Property1:} Importance \\
\textbf{Property2:} Responsibility \\
\textbf{Short Explanation:} In Iraq, legislators require more compliance with legal rules, such as having only one nationality, than average public officials. \\
\textbf{Long Explanation:} The former Iraqi nationality law imposes strict regulations against Iraqis who hold dual nationality. It is prohibited for such individuals to act as public officials, which include various positions within the government. However, the law goes even further with its restrictions on individuals who seek to become legislators, one of the highest positions of power in government. This law clearly indicates the importance of loyalty to one's country and the responsibility of government officials to represent their nation without potential conflicts of interest from dual nationality. \\
\textbf{Alternative Long Explanation:} The former Iraqi nationality law prohibits Iraqis from holding dual nationality and acting as public officials. Public officials include various positions within the government, such as administrators, judges, and diplomats. Legislators, on the other hand, are a specific group of public officials who are responsible for creating and passing laws. Since legislators hold a higher position of power and responsibility than other public officials, it is even less likely that those with dual nationality are allowed to act as legislators. This law reflects the importance of loyalty to one's country and the need for government officials to represent their nation without potential conflicts of interest from dual nationality. \\

\noindent \textbf{Example2:} \\
\textbf{Sentence:} The men on trial are all accused of being accessories to the murder, but the authorities have still failed to find the triggerman, let alone identify the mastermind of the killing. \\
\textbf{Correlate:} find the triggerman \\
\textbf{Remnant:} identify the mastermind of the killing \\
\textbf{Likelihood:} Yes \\
\textbf{Property1:} Skills needed \\
\textbf{Property2:} Effort intellectual \\
\textbf{Short Explanation:} Given that locating the triggerman already requires significant investigative skill, identifying the mastermind of the killing is naturally more challenging, as it demands an even greater level of expertise and intellectual effort. \\
\textbf{Long Explanation:} When a crime is committed, there is often a person who actually commits the crime and another who directed this person. Because they are physically on the crime scene, it is often easier for the authorities to find the triggerman. More energy, time and criminal ingenuity is required to find the hidden person who directed this all, the mastermind. Alternative explanation: the more physically removed from a crime a responsible person is, the more energy is needed to find them. A mastermind is more physically removed from a crime than a triggerman. \\

\noindent \textbf{Example3:} \\
\textbf{Sentence:} About his newfound fame: ``You have to remember it's unusual for me to even have a job, let alone to have a success or get a prize. \\
\textbf{Correlate:} have a job \\
\textbf{Remnant:} to have a success or get a prize \\
\textbf{Likelihood:} Yes \\
\textbf{Property1:} Effort intellectual \\
\textbf{Property2:} Skills needed \\
\textbf{Short Explanation:} Achieving success or winning a prize in the job encompasses more intellectual effort than securing employment. \\
\textbf{Long Explanation:} A person needs skills to have a job and not lose that job. In order to additionally have success in that job, or win prizes in that job, they need even more skills. The person states that it's already surprising to even hold down a job, so how much more surprising is it to be successful. \\

\noindent \textbf{Example4:} \\
\textbf{Sentence:} A French ban on gathering any kind of ethnic data -- in the name of republican equality -- has long hobbled its ability to measure, let alone tackle discrimination. \\
\textbf{Correlate:} measure \\
\textbf{Remnant:} tackle \\
\textbf{Likelihood:} Yes \\
\textbf{Property1:} Typical sequence of actions \\
\textbf{Property2:} Information Accessibility \\
\textbf{Short Explanation:} Because tackling discrimination requires measurement first, a French ban on gathering any kind of ethnic data in the name of republican equality has long hindered its ability to address discrimination. \\
\textbf{Long Explanation:} If we cannot gather data, there are two things we cannot do: measure how much discrimination there is, and then put in place measures against discrimination. In the speaker's mind, dealing with discrimination is only possible after the first (measuring) is accomplished (and neither of the two are currently accomplished). Although there is a scenario where one would tackle discrimination without even measuring it. \\

\noindent \textbf{Example5:} \\
\textbf{Sentence:} Each proposal is budgeted to cost the government up to six billion dollars, a fraction of the 30 billion dollars surplus already accumulated over the past six fiscal years, let alone the record 26 billion dollar windfall that analysts expect the government to achieve this year. \\
\textbf{Correlate:} 30 billion dollars surplus already accumulated over the past six fiscal years \\
\textbf{Remnant:} the record 26 billion dollar windfall that analysts expect the government to achieve this year \\
\textbf{Likelihood:} No \\
\textbf{Property1:} Amount of \\
\textbf{Property2:} Income \\
\textbf{Short Explanation:} The cost of the proposal is a fraction of the past six fiscal years' \$30 billion surplus, which pales in comparison to the unprecedented \$30 billion plus additional \$26 billion windfall that analysts predict for the government this year. \\
\textbf{Long Explanation:} The government is able to afford each proposal costing up to 6 billion dollars, since if it is only a fraction of the 30 billion dollars surplus accumulated over the past six fiscal years, it would be an even smaller fraction of the 30 billion dollars surplus accumulated over past six fiscal years the record 26 billion dollar windfall that analysts expect the government to achieve this year. \\

\chapter{Supplementary Evaluation Results}

\section{Assessing GPT-3.5-turbo's Comprehension of \textit{a fortiori} Logic} \label{appendix:understand_afortiori}
\begin{figure}[h]
    \centering
    \includegraphics[width=0.8\textwidth]{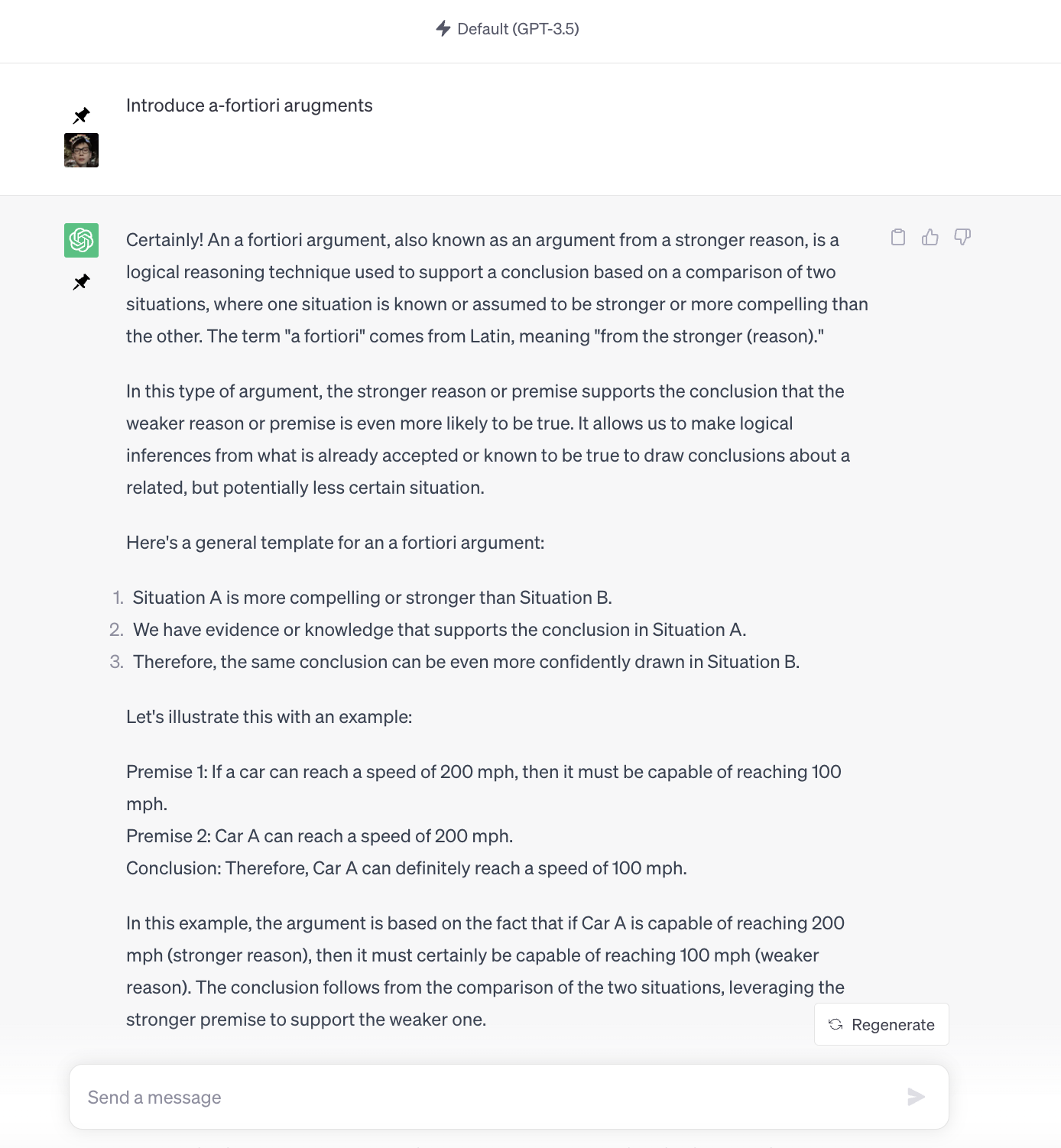}
    \caption{GPT-3.5-turbo demonstrates a correct reasoning path for \textit{a fortiori} arugments}
    \label{fig:understand_afortiori}
\end{figure}

\section{Comprehensive Statistical Analysis for Correlate and Remnant Extraction} \label{Appendix:stat_analysis}

\begin{table}[ht]
\centering
\begin{tabularx}{\textwidth}{|X|X|X|X|X|}
\hline
\textbf{Statistic} & \textbf{Correlate Similarity} & \textbf{Remnant Similarity} & \textbf{Correlate Exact Matches} & \textbf{Remnant Exact Matches} \\
\hline
Mean & 0.7217 & 0.8430 & 0.7088 & 0.8319 \\
\hline
Median & 0.8473 & 1.0000 & 0.8750 & 1.0000 \\
\hline
Standard Deviation & 0.3164 & 0.2654 & 0.3440 & 0.2934 \\
\hline
Minimum & -0.0132 & -0.0434 & 0.0000 & 0.0000 \\
\hline
25th Percentile & 0.4637 & 0.7171 & 0.4444 & 0.7330 \\
\hline
75th Percentile & 1.0000 & 1.0000 & 1.0000 & 1.0000 \\
\hline
Maximum & 1.0000 & 1.0000 & 1.0000 & 1.0000 \\
\hline
\end{tabularx}
\caption{Comprehensive statistical analysis of correlate and remnant extraction across the entire dataset (1,030 instances). This table presents the complete results, complementing the simplified version provided in the main content.}
\label{tab:stat_corre&rem_full}
\end{table}

\begin{table}[ht]
\centering
\begin{tabularx}{\textwidth}{|X|X|X|X|X|}
\hline
\textbf{Statistic} & \textbf{Correlate Similarity} & \textbf{Remnant Similarity} & \textbf{Correlate Exact Matches} & \textbf{Remnant Exact Matches} \\
\hline
Mean & 0.7163 & 0.8414 & 0.7049 & 0.8321 \\
\hline
Median & 0.8366 & 1.0000 & 0.8571 & 1.0000 \\
\hline
Standard Deviation & 0.3177 & 0.2664 & 0.3432 & 0.2920 \\
\hline
Minimum & -0.0132 & -0.0434 & 0.0000 & 0.0000 \\
\hline
25th Percentile & 0.4592 & 0.7101 & 0.4229 & 0.7273 \\
\hline
75th Percentile & 1.0000 & 1.0000 & 1.0000 & 1.0000 \\
\hline
Maximum & 1.0000 & 1.0000 & 1.0000 & 1.0000 \\
\hline
\end{tabularx}
\caption{Statistical analysis for correlate \& remnant extraction specifically for \textit{a fortiori} arguments (966 instances). This table presents the detailed results, complementing the condensed version given in the main content.}
\label{tab:stat_corre&rem_af}
\end{table}

\section{Experiment Results for Sentence Type and Logic Category Classification} \label{Appendix:Sent_logic}

\begin{figure}[ht]
    \centering
    \includegraphics[width=0.7\textwidth]{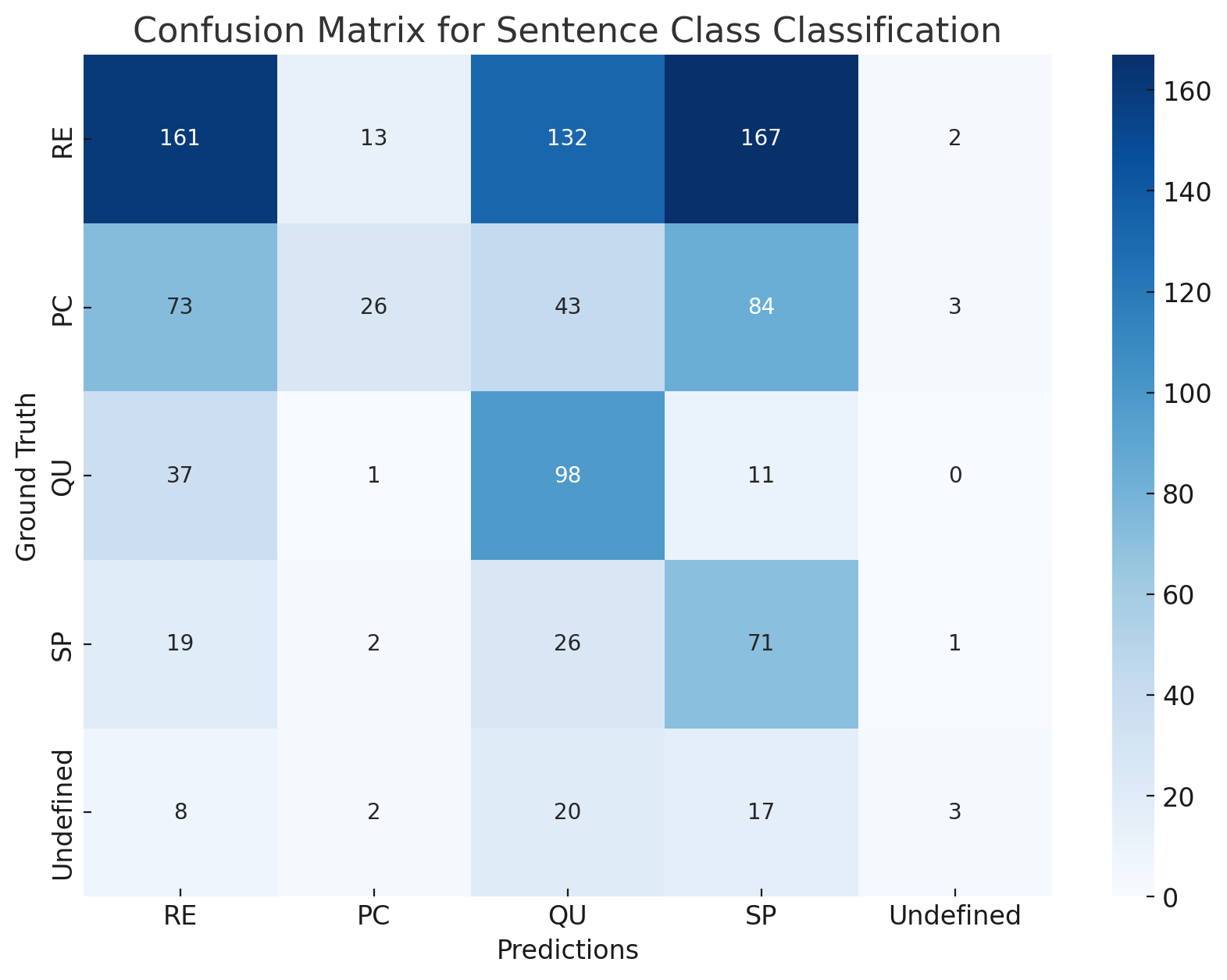}
    \caption{Confusion Matrix for Sentence Type Classification}
    \label{fig:confusion_sent}
\end{figure}

\begin{figure}[ht]
    \centering
    \includegraphics[width=0.7\textwidth]{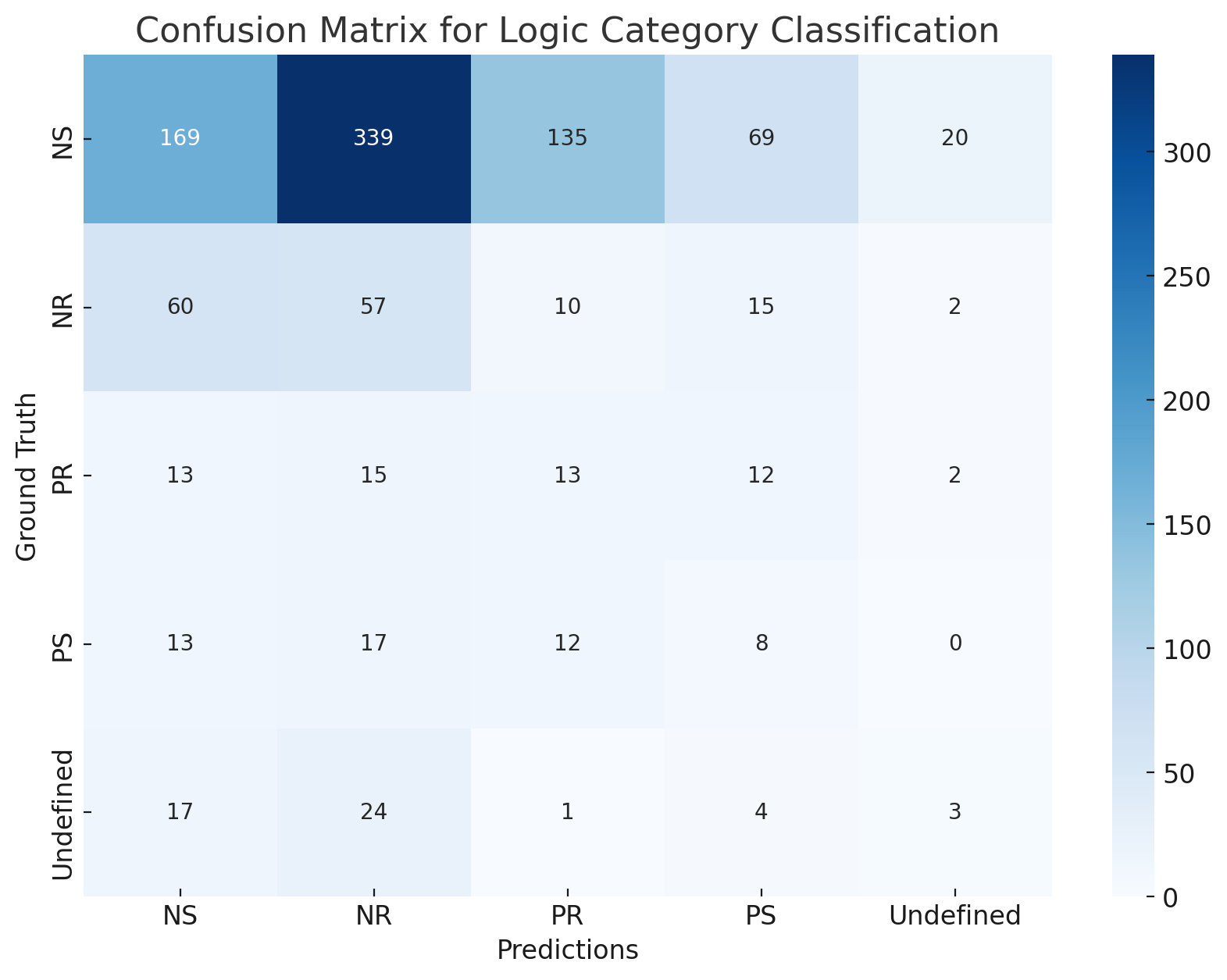}
    \caption{Confusion Matrix for Logic Category Classification}
    \label{fig:confusion_logic}
\end{figure}

\section{Top 3 Properties in Human Annotation and Model Prediction for each Sentence Type} \label{Appendix:top3property}

\begin{table}[ht]
    \centering
    \begin{tabular}{|p{2cm}|p{6.5cm}|p{5.5cm}|}
    \hline
    Sentence Type & Top Three Properties in Human Annotation & Top Three Properties in Model Prediction \\
    \hline
    RE            & 1. Typical sequence of actions (51)   & 1. Importance (67)                           \\
                  & 2. Financial cost (43)               & 2. Complexity (41)                           \\
                  & 3. Skills needed (35)                 & 3. Difficulty (37)                           \\
    SP            & 1. Skills needed (74)                 & 1. Specificity (221)                         \\
                  & 2. Specific case of (56)              & 2. Difficulty (31)                           \\
                  & 3. Typical sequence of actions (52)   & 3. Importance (21)                           \\
    QU            & 1. Skills needed (60)                 & 1. Quantity (54)                             \\
                  & 2. Number of (47)                     & 2. Importance (37)                           \\
                  & 3. Amount of (29)                     & 3. Difficulty (31)                           \\
    PC            & 1. Typical sequence of actions (19)   & 1. Complexity (4)                            \\
                  & 2. Effort intellectual (7)            & 2. Precondition (3)                          \\
                  & 3. Control level (6)                  & 3. Timing (3)                                \\
    \hline
    \end{tabular}
    \caption{Top 3 Properties in Human Annotation and Model Prediction for each Sentence Type when External Information is NOT provided. The numbers in the brackets refer to their frequencies}
    \label{tab:top3_property_woinfo}
\end{table}

\begin{table}[ht]
    \centering
    \begin{tabular}{|p{2cm}|p{6.5cm}|p{6.5cm}|}
    \hline
    Sentence Type & Top Three Properties in Human Annotation & Top Three Properties in Model Prediction \\
    \hline
    RE            & 1. Skills needed (111)                 & 1. Skills needed (103)                      \\
                  & 2. Control level (54)                   & 2. Control level (43)                       \\
                  & 3. Financial cost (44)                   & 3. Effort intellectual (39)                \\
    SP            & 1. Specific case of (82)                & 1. Specific case of (50)                    \\
                  & 2. Importance (9)                       & 2. Importance (13)                          \\
                  & 3. Typical sequence of actions (8)       & 3. Specificity (7)                          \\
    QU            & 1. Number of (53)                       & 1. Number of (36)                           \\
                  & 2. Amount of (36)                       & 2. Amount of (27)                           \\
                  & 3. Difficulty (16)                      & 3. Difficulty (18)                          \\
    PC            & 1. Typical sequence of actions (127)     & 1. Typical sequence of actions (89)          \\
                  & 2. Skills needed (30)                   & 2. Skills needed (25)                       \\
                  & 3. Effort intellectual (21)              & 3. Effort intellectual (22)                \\
    \hline
    \end{tabular}
    \caption{Top 3 Properties in Human Annotation and Model Prediction for each Sentence Type when External Information is provided. The numbers in the brackets refer to their frequencies}
    \label{tab:top3_property_winfo}
\end{table}

\section{Top 10 Common properties shared by human annotation and model prediction, ranked by their frequencies}

\begin{table}[ht]
\begin{center}
\begin{tabular}{|c|l|p{4cm}|p{4.5cm}|}
\hline
\textbf{Rank} & \textbf{Property} & \textbf{Frequency in Human Annotation} & \textbf{Frequency in Model Prediction} \\
\hline
1 & Skills needed & 176 & 130 \\
2 & Effort intellectual & 74 & 64 \\
3 & Information accessibility & 61 & 34 \\
4 & Number of & 60 & 25 \\
5 & Proximity & 55 & 20 \\
6 & Familiarity & 45 & 20 \\
7 & Amount of & 43 & 17 \\
8 & Importance & 39 & 13 \\
9 & Effort physical & 38 & 12 \\
10 & Danger & 36 & 9 \\
\hline
\end{tabular}
\end{center}
\caption{Top 10 common properties shared by human annotation and model prediction without external information}
\label{tab:top10commen_woinfo}
\end{table}

\begin{table}[ht]
\begin{center}
\begin{tabular}{|c|l|p{4cm}|p{4.5cm}|}
\hline
\textbf{Rank} & \textbf{Property} & \textbf{Frequency in Human Annotation} & \textbf{Frequency in Model Prediction} \\
\hline
1 & Skills needed & 180 & 166 \\
2 & Typical sequence of actions & 159 & 146 \\
3 & Specific case of & 98 & 79 \\
4 & Control level & 78 & 75 \\
5 & Financial cost & 76 & 70 \\
6 & Effort intellectual & 74 & 67 \\
7 & Information accessibility & 61 & 65 \\
8 & Number of & 60 & 63 \\
9 & Effort emotional & 56 & 56 \\
10 & Proximity & 56 & 54 \\
\hline
\end{tabular}
\end{center}
\caption{Top 10 common properties shared by human annotation and model prediction with external information}
\label{tab:top10commen_withinfo}
\end{table}

\section{Grammar Errors in Original Sentences, Generated Explanations, and Augmented Sentences} \label{Appendix:grammar_error}

\begin{table}[ht]
\centering
\begin{tabular}{|p{5cm}|p{2cm}|p{6cm}|}
\toprule
\textbf{Sources} & \textbf{Number of Entries with Errors} & \textbf{Common Error Types} \\
\hline
Original Sentence & 55 & Spelling: 45, Whitespace: 3, Apostrophe: 1 \\
\hline
short\_explain\_with all external information & 117 & Spelling: 53, Whitespace: 14, Capitalization: 8 \\
\hline
long\_explain\_with all external information & 193 & Spelling: 103, Whitespace: 19, Capitalization: 16 \\
\hline
short\_explain\_without external information & 91 & Spelling: 40, Whitespace: 9, Capitalization: 6 \\
\hline
long\_explain\_without external information & 179 & Spelling: 85, Whitespace: 12, Capitalization: 12 \\
\bottomrule
\end{tabular}
\caption{Grammar Errors in Original Sentences and their Explanations, detected by LanguageTool \cite{LanguageTool}}
\label{tab:combined_error_info}
\end{table}

\begin{table}[ht]
\centering
\begin{tabular}{|p{5cm}|p{3cm}|p{6cm}|}
\toprule
\textbf{Augmentation Strategies} & \textbf{Number of Entries with Errors} & \textbf{Common Error Types} \\
\hline
Similar Semantic Sentences & 63 & Uppercase\_sentence\_start: 18, Morfologik\_rule\_en\_us: 17, Comma\_compound\_sentence: 8 \\
\hline
Novel Sentences & 163 & Morfologik\_rule\_en\_us: 126, Uppercase\_sentence\_start: 19, Comma\_compound\_sentence: 7 \\
\bottomrule
\end{tabular}
\caption{Grammar Errors in Original Sentences and Augmented Sentences, detected by LanguageTool \cite{LanguageTool}}
\label{tab:aug_error_info}
\end{table}

\section{Comprehensive Statistic Analysis of Semantic Similarity for Augmented Sentences} \label{Appendix:stat_analysis_aug}

\begin{table}[h]
\centering
\begin{tabular}{lcc}
\hline
 & Augmented Similar Sentences & Augmented Novel Sentences \\
\hline
Mean & 0.5956 & 0.3012 \\
Standard Deviation & 0.2634 & 0.1421 \\
Minimum & -0.0193 & -0.0645 \\
25th Percentile (Q1) & 0.3653 & 0.1974 \\
Median (Q2) & 0.6258 & 0.2917 \\
75th Percentile (Q3) & 0.8315 & 0.4022 \\
Maximum & 0.9959 & 0.8176 \\
\hline
\end{tabular}
\caption{Statistic Analysis of Semantic Similarity Scores for Augmented Sentences}
\label{tab:stat_semantic_aug}
\end{table}

\section{Top 10 Frequent Topics for Original and Augmented Sentences}

\begin{figure}[h!]
\centering
\includegraphics[width=1\textwidth]{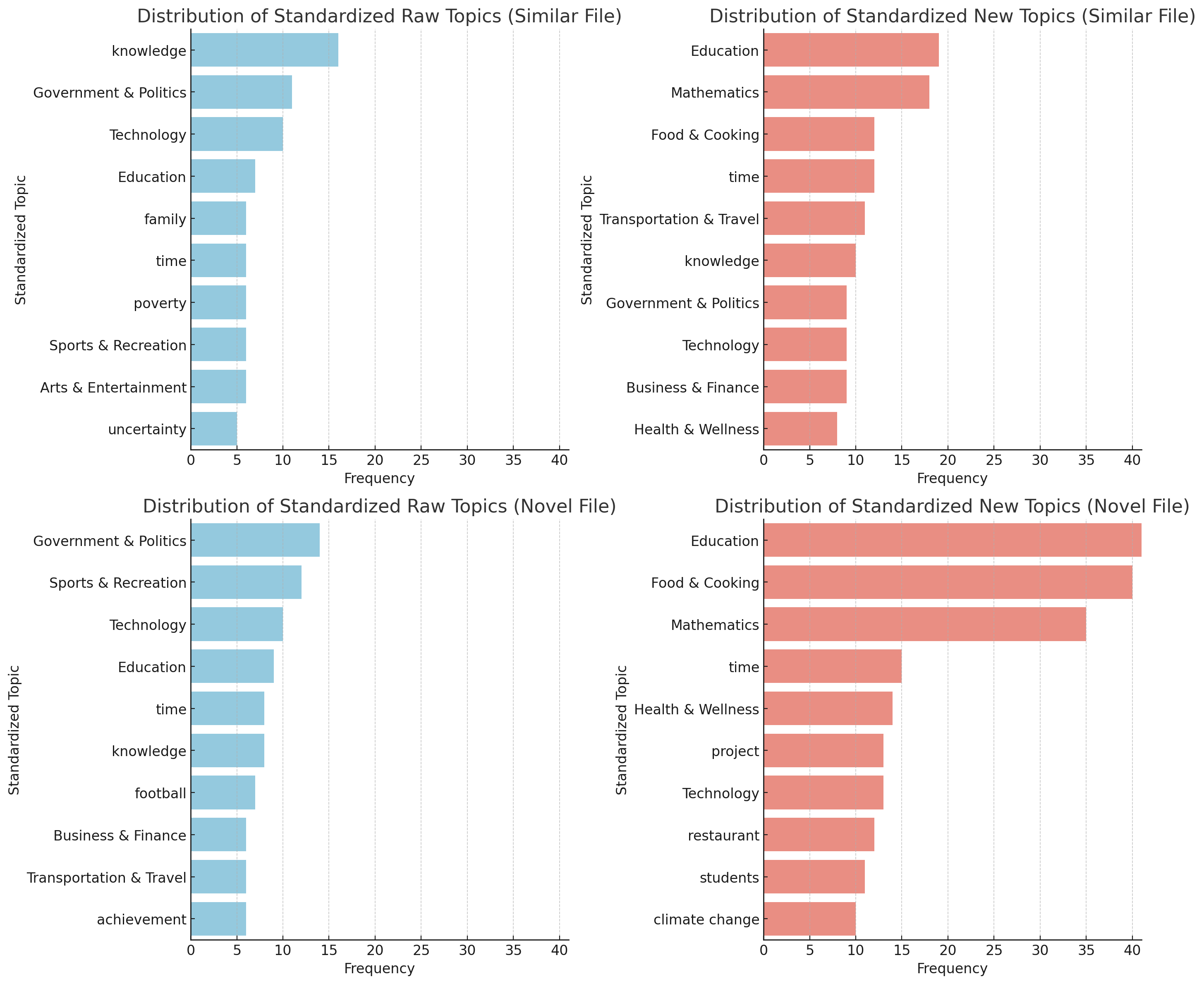}
\caption{Top 10 Topics for Original and Augmented Sentences}
\label{fig:common_topic}
\end{figure}

\label{lastpage}
\end{document}